\documentclass[10pt,twocolumn,letterpaper]{article}

\usepackage{iccv}
\usepackage{times}
\usepackage{epsfig}
\usepackage{graphicx}
\usepackage{amsmath}
\usepackage{amssymb}

\usepackage{caption}
\usepackage{booktabs}
\usepackage{footnote}
\usepackage{cuted}
\usepackage{multirow}
\usepackage[square,comma,sort&compress,numbers]{natbib}
\usepackage[accsupp]{axessibility}

\usepackage[pagebackref=true,breaklinks=true,letterpaper=true,colorlinks,bookmarks=false]{hyperref}

\iccvfinalcopy % *** Uncomment this line for the final submission

\def\iccvPaperID{4636} % *** Enter the ICCV Paper ID here

\ificcvfinal\fi

\begin{document}

%%%%%%%%% TITLE
\title{GRAM-HD: 3D-Consistent Image Generation at High Resolution with Generative Radiance Manifolds}

\author{Jianfeng Xiang\thanks{Work done when JX and YD were interns at MSRA.}\,\,$^{1,2}$ \quad Jiaolong Yang$^{2}$ \quad Yu Deng$^{*1,2}$ \quad Xin Tong$^{2}$ \\
	$^1${Tsinghua University} \quad $^2${Microsoft Research Asia} \\
    {\tt\small \{t-jxiang,jiaoyan,xtong\}@microsoft.com \quad  dengyu2008@hotmail.com}
}

\maketitle
\ificcvfinal\thispagestyle{empty}\fi

%%%%%%%%% TEASER
\begin{strip}
    \vspace{-57pt}
    \centering
    \captionsetup{type=figure}
    \includegraphics[width=1.0\textwidth]{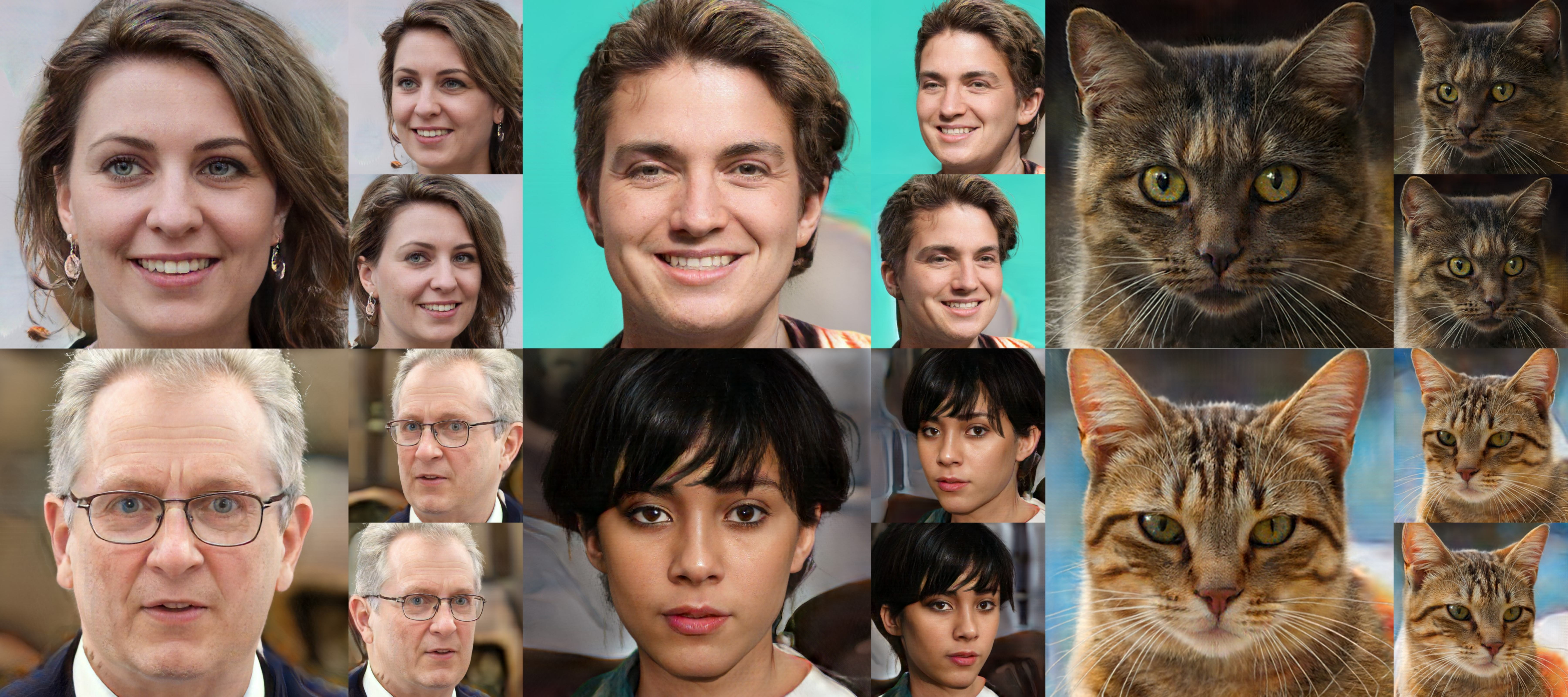}
    \vspace{-20pt}
	\caption{Curated generation samples of 1024$^2$ human faces 512$^2$ cats from GRAM-HD. Our method can generate high-resolution, high-quality and strict 3D-consistent images with explicitly controllable poses. Note the realistic geometry details such as the whisker of cats. (\textbf{Best viewed with zoom-in; see also the \href{https://jeffreyxiang.github.io/GRAM-HD/}{\emph{project page}} for more results with videos.})}	\label{fig:results}
\end{strip}

\footnotetext[1]{Project page: \href{https://jeffreyxiang.github.io/GRAM-HD/}{https://jeffreyxiang.github.io/GRAM-HD/}}

%%%%%%%%% ABSTRACT
\begin{abstract}
   \vspace{-12pt}
   Recent works have shown that 3D-aware GANs trained on unstructured single image collections can generate multiview images of novel instances. The key underpinnings to achieve this are a 3D radiance field generator and a volume rendering process. However, existing methods either cannot generate high-resolution images (e.g., up to 256$\times$256) due to the high computation cost of neural volume rendering, or rely on 2D CNNs for image-space upsampling which jeopardizes the 3D consistency across different views. This paper proposes a novel 3D-aware GAN that can generate high resolution images (up to 1024$\times$1024) while keeping strict 3D consistency as in volume rendering. Our motivation is to achieve super-resolution directly in the 3D space to preserve 3D consistency. We avoid the otherwise prohibitively-expensive computation cost by applying 2D convolutions on a set of 2D radiance manifolds defined in the recent generative radiance manifold (GRAM) approach, and apply dedicated loss functions for effective GAN training at high resolution. Experiments on FFHQ and AFHQv2 datasets show that our method can produce high-quality 3D-consistent results that significantly outperform existing methods. It makes a significant step towards closing the gap between traditional 2D image generation and 3D-consistent free-view generation.
   \footnotemark[1]
   \vspace{-12pt}
\end{abstract}

%%%%%%%%% BODY TEXT
\section{Introduction}

\label{sec:intro}

While generative modeling of 2D images have achieved tremendous success~\cite{brock2018large,karras2019style,karras2020analyzing,karras2021alias} with generative adversarial networks (GANs)~\cite{goodfellow2014generative}, 3D-aware GANs that aims to generate photorealistic multiview images begun to emerge in recent years~\cite{schwarz2020graf,niemeyer2021giraffe,chan2021pi,deng2021gram, skorokhodov2022epigraf, zhao2022gmpi}. Despite both being trained on unstructured 2D image collections, the latter is capable of synthesizing the images of an object at different 3D viewpoints. The key to achieve this is to generate an underlying 3D representation, for which neural radiance field (NeRF)~\cite{mildenhall2020nerf} has been the cornerstone for recent methods. With volumetric rendering, NeRF can produce realistic images while enforcing strong 3D consistency across views.

However, the high computational cost of neural volumetric rendering greatly limits the affordable image resolution for GAN training. It also introduces hurdles in fine detail generation due to insufficient point sampling. A workaround is only judging whether a patch of the generated image is real to not during training~\cite{schwarz2020graf, skorokhodov2022epigraf} rather than a whole image. But using a patch discriminator may lack global perception of the images and lead to inferior image generation quality. The recent generative radiance manifold (GRAM) method~\cite{deng2021gram} significantly improved the generation quality by sampling points on a set of learned surface manifolds. Still, it can only be trained on images up to 256$\times$256 resolution on modern GPUs, which is in sheer contrast to state-of-the-art 2D GANs that can easily model 1024$\times$1024 images with moderate computing cost.

Along a different axis, many methods resort to 2D convolutions to tackle the dilemma \cite{gu2021stylenerf,niemeyer2021giraffe,or2022stylesdf, zhang2022multi,chan2022efficient}. A straightforward idea shared by these methods is to render low-resolution images or feature maps and apply 2D CNNs to increase the resolution. With this strategy, they have demonstrated higher-resolution generation (\eg, 512$\times$512 for \cite{zhang2022multi,chan2022efficient} and 1024$\times$1024 for \cite{gu2021stylenerf,or2022stylesdf}). Unfortunately, image-space upsampling with 2D CNNs inevitably incurs 3D inconsistency among the generated multiview images. As such, these methods can be used in user-interactive image generation and manipulation but are not suitable for video synthesis and animation. 

We propose \emph{GRAM-HD}, a GAN method that can synthesize strongly 3D-consistent images at high resolution. Our motivation is to do upsampling or super-resolution in the 3D space and keep the volume rendering paradigm to retain strict 3D consistency. But how to achieve this efficiently is not straightforward (\eg, upsampling a discretized low-resolution volume using 3D convolutions quickly becomes untractable for high-resolution output). In this paper, we leverage the GRAM~\cite{deng2021gram} method, which defines a set of surface manifolds, to handle high resolution generation. Our key insight is that the surface manifolds can be upsampled using 2D CNNs for efficient super-resolution.
We flatten and sample each learned surface to regular 2D image grids and apply a shared 2D CNN for upsampling and feature-to-radiance translation. This way, our method not only ensures multiview consistency by generating a high-resolution 3D representation for rendering, but also enjoys the computational efficiency of 2D CNNs. In essence, \emph{we tackle a 3D super-resolution task with object-space 2D CNN}.

We evaluate our method on the FFHQ~\cite{karras2019style} and AFHQv2-CATS~\cite{choi2020stargan} datasets. We show that GRAM-HD can generate photorealistic images that are both of high resolution (up to 1024$\times$1024) and strongly multiview-consistent, which cannot be achieved by any previous method. It also outperforms GRAM in terms of both generation quality and speed at the same image resolution by upsampling low-resolution manifolds. 

\vspace{2pt}
\textbf{The contributions of this work} are summarized below:
\begin{itemize}
	\vspace{-5pt}
	\item We present a novel 3D-aware image generation approach that can generate high-resolution images (up to 1024$\times$1024) with strong multiview-consistency and highly-realistic geometry details.
	\vspace{-5pt}
	\item We introduce a method for 3D space super-resolution using efficient 2D CNN under the radiance manifold representation.
	\vspace{-5pt}
	\item We significantly reduced the computation cost of the  radiance manifold based 3D-aware generation method while obtaining higher quality images (\eg, 76\%$\downarrow$ memory cost, 58\%$\downarrow$ inference time, 21\%$\downarrow$ FID on FFHQ-$256^2$; 95\%$\downarrow$ inference time for $1024^2$)
\end{itemize}

\section{Related Work}

\paragraph{Neural scene representations} Neural scene representations~\cite{dosovitskiy2016learning,tatarchenko2016multi,eslami2018neural,sitzmann2019deepvoxels,park2019deepsdf,mescheder2019occupancy,niemeyer2020differentiable,sitzmann2019scene,sitzmann2020implicit,mildenhall2020nerf,wang2021neus,oechsle2021unisurf,sitzmann2021light} have seen tremendous progress in the past several years and shown promising results for modeling complex 3D scenes. Among these methods, NeRF~\cite{mildenhall2020nerf} and its variants~\cite{martin2021nerf,barron2021mip,barron2022mip} excel at learning detailed scene structures from a collection of posed images and synthesizing 3D-consistent novel views. Nevertheless, the high computational complexity of NeRF restricts its power when applied to reconstruction of 
numerous instances~\cite{jang2021codenerf,liu2021editing} or generative modeling of object categories~\cite{schwarz2020graf,chan2021pi}. Several methods have been proposed to reduce the computation cost of NeRF, including utilizing sparse data structures~\cite{liu2020neural,yu2021plenoxels,hedman2021baking},
introducing multi-resolution feature encoding~\cite{martel2021acorn,muller2022instant}, or reducing sampling points during rendering~\cite{lindell2021autoint,sitzmann2021light}. However, it remains unclear how these approaches can be applied to a generative modeling paradigm. Very recently, GRAM~\cite{deng2021gram} proposes to regulate radiance field learning on 2D manifolds and prominently improves the image generation quality at a resolution of $256\times256$. This paper inherits the manifolds representation of GRAM, and extends it to 3D-consistent image generation of high resolution. 

\vspace{-12pt}
\paragraph{Generative 3D-aware image synthesis} 3D-aware generative models~\cite{nguyen2019hologan,liao2020towards,shi2021lifting,schwarz2020graf,chan2021pi,niemeyer2021giraffe} aim to learn multiview image synthesis of an object category given uncontrolled 2D images collections. They achieve this by incorporating 3D representation learning into GAN training. Earlier works~\cite{nguyen2019hologan,nguyen2020blockgan,szabo2019unsupervised} utilize mesh or voxel to represent the underlying 3D scenes. Recently, plenty of works~\cite{schwarz2020graf,chan2021pi,xu2021generative,pan2021shading} leverage NeRF's volumetric representation to achieve image generation with higher 3D-consistency. However, NeRF's high computational cost prevents them from generating high-resolution images with fine details. Most of the follow-up works~\cite{niemeyer2021giraffe,gu2021stylenerf,zhou2021cips,chan2022efficient,or2022stylesdf,zhang2022multi,xu20213d} handle this problem by first rendering low-resolution images or feature maps and then applying 2D CNNs for super-resolution. They suffer from a common 3D inconsistency issue when varying camera viewpoints, due to the black-box rendering of CNNs. Some methods~\cite{gu2021stylenerf,chan2022efficient} alleviate this problem by enforcing constrains between the generated high-resolution images and their low-resolution counterparts, 
but still cannot guarantee the consistency of high-frequency details. Different from the above methods, \cite{deng2021gram} learns radiance manifolds and applies manifold rendering~\cite{deng2021gram,zhou2018stereo} to achieve high-quality and multiview-consistent image synthesis, but it has difficulties to generate images with a resolution beyond $256\times256$. In this paper, we propose a novel approach to learn high-resolution radiance manifolds, and achieve strongly multiview-consistent image generation at high resolution.

\vspace{-12pt}
\paragraph{Image super-resolution} Image super-resolution is a longstanding task in computer vision and has seen enormous developments~\cite{chang2004super,glasner2009super,dong2015image,ledig2017photo,lim2017enhanced,zhang2018image,wang2018esrgan,zhang2020deep,chen2021learning,wang2021real}. We leverage the techniques from this field, but for super-resolution of radiance manifolds in 3D space instead of 2D images.

\section{Approach}

\begin{figure*}[t]
	\centering
	\includegraphics[width=1\textwidth]{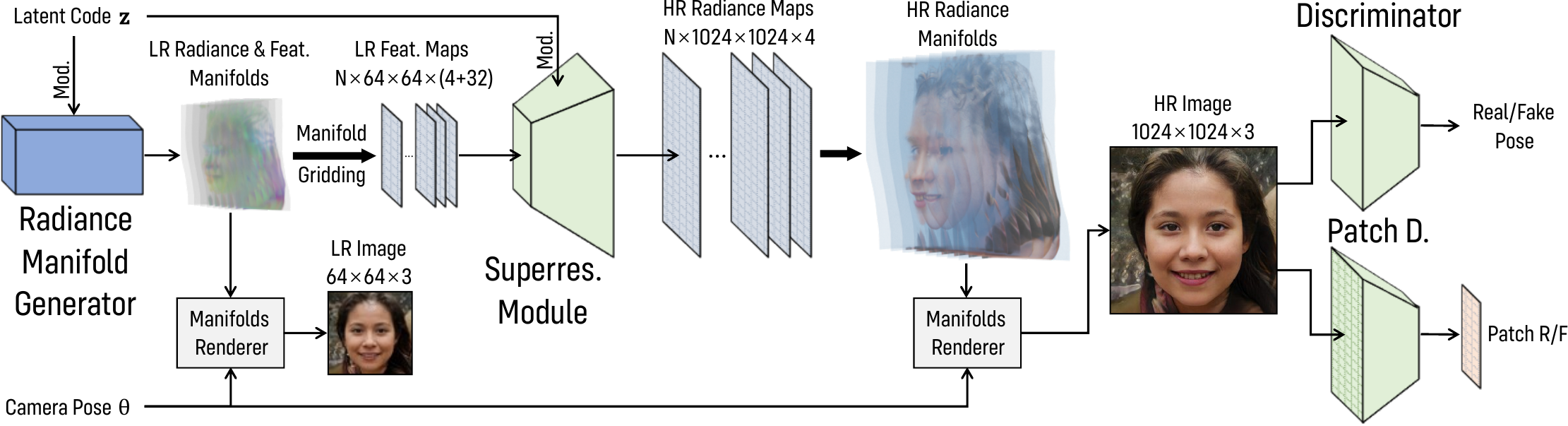}
	\vspace{-19pt}
	\caption{The overall framework of our GRAM-HD method. The generator consists of two components: the radiance manifold generator and the manifold super-resolution module. The former generates radiance and feature manifolds that represent an LR 3D scene. Through manifold gridding, the manifolds are sampled to discrete 2D feature maps. The super-resolution module then processes these feature maps and output HR radiance maps. Finally, an HR image can rendered by computing ray-manifold intersections and integrating their radiance sampled from the HR radiance maps. Note that \emph{not like} previous works utilizing \emph{2D image super-resolution} for HR image generation, we directly do \emph{3D representation super-resolution} and keep the volume rendering paradigm, thus keep the strong 3D consistency of the output images.}
	\label{fig:framework}
 \vspace{-8pt}
\end{figure*}

Given a collection of 2D images, our method aims to learn, through adversarial learning, a 3D-aware image generator $G$ which takes a latent code $\boldsymbol z\in \mathbb{R}^{d_z}\sim p_z$ and an explicit camera pose $\boldsymbol \theta\in\mathbb{R}^3\sim p_\theta$ as inputs, and outputs a synthesized image $I$ of an virtual instance determined by $\boldsymbol z$ under pose $\boldsymbol\theta$:

\vspace{-6pt}\begin{equation}
	G:(\boldsymbol z, \boldsymbol \theta)\in\mathbb R ^{d_z+3}\rightarrow I\in\mathbb R^{H\times W\times 3}.
\vspace{-0pt}\end{equation}

Figure~\ref{fig:framework} depicts the overall framework of our method. We first use a radiance manifold generator to generate the radiance and feature manifolds representing a low-resolution scene. Then we flatten and discretize the manifolds and upsample them using a shared CNN to get an high-resolution representation. After training, GRAM-HD can render high-definition and 3D-consistent images. 

\subsection{Radiance Manifolds}

The concept of radiance manifolds is introduced in GRAM~\cite{deng2021gram} and we briefly review it here. The original neural radiance filed~\cite{mildenhall2020nerf} models the radiance of a continuous 3D scene with a neural network, and volumetric rendering is done by sampling points along each viewing ray and integrating the radiance. However, the high memory and computation cost greatly restricts the number of point samples for GAN training. The GRAM method regulates point sampling and
radiance field learning on a set of learned surface manifolds which can significantly improve image generation quality.

The manifolds are embodied as a set of iso-surfaces in a 3D scalar field. This scalar field is shared for all generated instances and represented by a light-weight MLP called the \emph{manifold predictor} $\mathcal M$:
\vspace{-3pt}\begin{equation}
	\mathcal M: \boldsymbol x\in\mathbb R^3\rightarrow s\in\mathbb R.
\vspace{-3pt}\end{equation}
With $\mathcal M$, the iso-surfaces $\{S_i\}$ can be extracted with $N$ predefined levels $\{l_i\}$: 
\vspace{-3pt}\begin{equation}
	S_i=\{\boldsymbol x|\mathcal M(\boldsymbol x)=l_i \}.
\vspace{-3pt}\end{equation}
For rendering, point samples are calculated as the intersections between each viewing ray $\boldsymbol r$ and the extracted surfaces:
\vspace{-6pt}\begin{equation}
	\{\boldsymbol x_{i}\}=\{\boldsymbol x|\boldsymbol x\in \boldsymbol r\cap \{S_j\}\}.
\vspace{-0pt}\end{equation}
Radiance of these point samples are then generated by the \emph{radiance generator} $\Phi$, which is an MLP modulated by latent code $\boldsymbol z$:
\vspace{-3pt}\begin{equation}
	\Phi:(\boldsymbol z,\boldsymbol x,\boldsymbol d)\in\mathbb{R}^{d_z+6}\rightarrow(\boldsymbol c, \alpha)\in\mathbb R^4,
\vspace{-3pt}\end{equation}
where $\boldsymbol d$ is a view direction, $\boldsymbol c$ is the color, and $\alpha$ is the occupancy. The final color value for each ray can be computed with the rendering equation~\cite{oechsle2021unisurf,zhou2018stereo}:
\vspace{-3pt}\begin{equation}
	C({\boldsymbol r})
	=   \sum_{i=1}^{N}\prod_{j<i}(1-\alpha({\boldsymbol x}_j))\alpha({\boldsymbol x}_i)\boldsymbol c({\boldsymbol x}_i). \label{eq:render}
\vspace{-3pt}\end{equation}
Note the latent code $\boldsymbol z$ and view direction $\boldsymbol d$ are omitted here for brevity.

Our radiance manifolds are adapted from GRAM with a few twists. First, we only use these MLP-generated radiance manifolds to represent a coarse, low-resolution scene trained to render 64$\times$64 images. Second, since the intermediate feature also bears useful information, we concatenate them with the final output color and occupancy as the input to the subsequent manifold super-resolution.

\subsection{Manifold Super-Resolution}

Our key insight for manifold super-resolution is that efficient 2D CNNs can be applied. To this end, we first map the manifolds to regular image grids.

\vspace{-12pt}
\paragraph{Manifold gridding} The manifold gridding operation $\Pi$ flattens the learned surfaces $\{S_i\}$ and samples them to get a set of low-resolution radiance and feature maps:
\vspace{-3pt}\begin{equation}
	\Pi:\{S_i\} \rightarrow \{R_i\}\doteq \boldsymbol{R}_{\mathrm{lr}}\in \mathbb{R}^{N,H_{\mathrm{lr}},W_{\mathrm{lr}},4},
\vspace{-3pt}\end{equation}
where $H_{\mathrm{lr}}$ and $W_{\mathrm{lr}}$ are the spatial resolution for which we use 64$\times$64.
Such a gridding operation can be implemented in various ways. GRAM has shown that the learned surfaces are nearly planner. So in this work we simply use orthogonal projection to flatten the surfaces. 

Specifically, we sample the surfaces by shooting $H_{\mathrm{lr}}\times W_{\mathrm{lr}}$ rays
that are parallel to Z-axis.
For the first $N\!-\!1$ surfaces, the sampling area encompasses the foreground objects in 3D. For the last surface which is the background plane in GRAM, we use a larger area as it spans a much wider region for covering the background under extreme views. We also apply an additional nonlinear mapping to sample denser points around the center region (see \emph{suppl. material} for details). After concatenating the sampled color, occupancy, and intermediate features from $\Phi$, the manifold gridding produces a 4D tensor $\boldsymbol{R}_{\mathrm{lr}}^+\in \mathbb{R}^{N,H_{lr},W_{lr},4+d_f}$ where $d_f$ is the channel number of features.

\vspace{-12pt}
\paragraph{Super-resolution CNN} 
To upsample the LR radiance maps, we condition the super-resolution CNN $\mathcal{U}$ with latent code $\boldsymbol z$ such that the added details are also controlled by $\boldsymbol z$ to the extent possible:
\vspace{-3pt}\begin{equation}
	\mathcal{U}:\begin{gathered}
	    (\boldsymbol z\in\mathbb{R}^{d_z}, \boldsymbol{R}^+_\mathrm{lr}\in\mathbb{R}^{N,H_\mathrm{lr},W_\mathrm{lr},4+d_f})\\
	    \rightarrow \boldsymbol{R}_\mathrm{hr}\in\mathbb{R}^{N,H_\mathrm{hr},W_\mathrm{hr},4},
	\end{gathered}
\vspace{-3pt}\end{equation}
where $H_\mathrm{hr}$ and $W_\mathrm{hr}$ are the HR resolution. 
Considering that the distribution of background radiance significantly differ from the foreground, we employ \emph{two CNNs}, one for the first $N-1$ foreground maps and another much smaller one (with half channels of the former) for the last background map. Adding this dedicated background network leads to higher image quality as we will show in the experiment. See also the \emph{suppl. material} for more details.

Our CNN architecture is adapted from previous image super-resolution networks~\cite{wang2018esrgan,shi2016real, zhang2018residual}. Specifically, we first apply Residual-in-Residual Dense Block (RRDB)~\cite{wang2018esrgan} to process the LR radiance and feature maps, and then use sub-pixel convolutions~\cite{shi2016real, zhang2018residual} for upsampling. RRDB is a CNN equipped with massive residual and dense connections which can effectively process the features, while sub-pixel convolution is a learnable upsampling method proven to work well for super-resolution. We empirically found that their combination leads to better performance.

We inject style information encoded by $\boldmath z$ using the weight modulation method of \cite{karras2020analyzing}.
An MLP is employed to map $\boldsymbol z$ to $s_i$, and apply weight modulation to all convolutional layers after RRDB. More details of our CNN architecture can be found in the \emph{suppl. material}.

\vspace{-12pt}
\paragraph{Image rendering} 
To render the final image after obtaining the HR radiance maps, we first calculate the intersections between the view rays and surface manifolds as in \cite{deng2021gram}. Then we use the same mapping function as in manifold gridding to get their projected 2D positions on the radiance maps and obtain radiance via bilinear interpolation. Pixel colors can then be calculated per Eq.~\ref{eq:render}.

\subsection{Network Training}
To train our model, we randomly sample latent code $\boldsymbol z$, camera pose $\boldsymbol\theta$ and real image $I$ from prior distributions $p_z$, $p_\theta$, and $p_\mathrm{real}$. We first use the following two loss functions as in GRAM~\cite{deng2021gram}. 

\begin{figure*}[t]
\vspace{-2pt}
	\centering
	\includegraphics[width=0.922\textwidth]{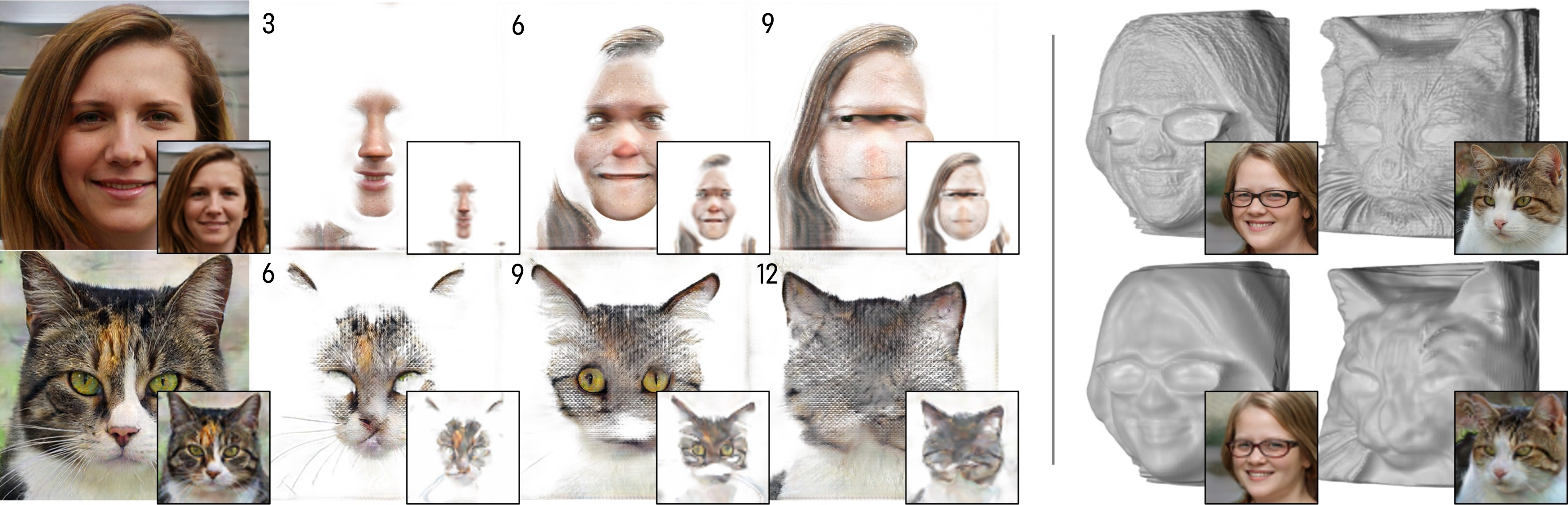}
	\vspace{-4pt}
	\caption{\textbf{Left:} Rendered images and radiance maps on the surface manifolds. Three sampled manifolds are shown here. The corresponding LR results before super-resolution are presented at bottom right. \textbf{Right:} Extracted proxy 3D shapes at HR (top row) and LR (bottom row). The rendered images are shown at bottom right for reference.}
	\label{fig:superres}
	\vspace{-8pt}
\end{figure*}

\vspace{-12pt}
\paragraph{Adversarial loss}
For adversarial learning~\cite{goodfellow2014generative}, a discriminator $D$ is used to distinguish the generated images from the real ones and compete with the generator $G$. We use the non-saturating GAN loss with R1 regularization~\cite{mescheder2018training} for training:
\vspace{-3pt}\begin{equation}\begin{split}
	\mathcal{L}_\mathrm{adv}=\ &\mathbb{E}_{\boldsymbol z\sim p_z,\boldsymbol \theta\sim p_\theta}[f(D(G(\boldsymbol z,\boldsymbol \theta)))]\\
	&+ \mathbb{E}_{I\sim p_\mathrm{real}}[f(-D(I))+\lambda\|\nabla D(I)\|^2],
\end{split}\vspace{-3pt}\end{equation}
where $f(x)=\log(1+\exp x)$ is the softplus function. Two discriminators are applied for the low and high resolution, respectively. The network architectures are similar to GRAM discriminator~\cite{deng2021gram}.

\vspace{-12pt}
\paragraph{Pose loss} Following GRAM~\cite{deng2021gram}, we add a loss to regularize the generate poses and ensure correct 3D geometry. Specifically, the discriminators also estimate a camera pose for each image, and $L_2$ loss is calculated between given camera pose and the estimated one:
\vspace{-3pt}\begin{equation}\begin{split}
	\mathcal{L}_\mathrm{pose}=\ &\mathbb{E}_{\boldsymbol z\sim p_z,\boldsymbol \theta\sim p_\theta}\|D^p(G(\boldsymbol z,\boldsymbol \theta))-\boldsymbol\theta\|^2\\
	&+ \mathbb{E}_{I\sim p_\mathrm{real}}\|D^p(I)-\hat{\boldsymbol\theta}\|^2.
\end{split}\vspace{-3pt}\end{equation}

We also introduce two new losses below, which are designed for our high resolution generation task.

\vspace{-12pt}
\paragraph{Patch adversarial loss}
We apply an extra patch discriminator~\cite{zhu2017unpaired} to the generated final images, which can help remove the checkerboard artifacts arisen due to upsampling. The patch adversarial loss can be written as:
\begin{equation}\begin{split}
	\mathcal{L}_\mathrm{patch}=\ &\mathbb{E}_{\boldsymbol z\sim p_z,\boldsymbol \theta\sim p_\theta}[f(D_\mathrm{patch}(G(\boldsymbol z,\boldsymbol \theta)))]\\
	&+ \mathbb{E}_{I\sim p_\mathrm{real}}[f(-D_\mathrm{patch}(I))]
\end{split}\end{equation}

\vspace{-12pt}
\paragraph{Cross-resolution consistency loss} Finally, we enforce the consistency between the generated LR and HR contents. We apply an $L_2$ difference to penalty the difference in both the rendered images and radiance maps:
\vspace{-3pt}\begin{equation}\begin{split}
	\mathcal{L}_\mathrm{cons}=\mathbb{E}_{\boldsymbol z\sim p_z,\boldsymbol \theta\sim p_\theta}[&\|\Gamma(G_\mathrm{hr}(\boldsymbol z,\boldsymbol \theta))-G_\mathrm{lr}(\boldsymbol z,\boldsymbol \theta)\|^2\\
	&+ \|\Gamma(\boldsymbol R_\mathrm{hr})-\boldsymbol R_\mathrm{lr}\|^2],
\end{split}\vspace{-3pt}\end{equation}
where $\Gamma$ denotes the bicubic downsampling operator. This loss enforces the super-resolution module to focus on fine detail generation without introducing dramatic change.

We employ a two-stage training strategy in our implementation. At the first stage, we train $G_\mathrm{lr}$ and $D_\mathrm{lr}$ (\emph{i.e.}, the GRAM model) for low-resolution image generation using $\mathcal{L}_\mathrm{adv}$ and $\mathcal{L}_\mathrm{pose}$. Then we froze $G_\mathrm{lr}$ and train high resolution generation using all the four loss functions.

\section{Experiments}\label{sec:experiments}

\paragraph{Implementation details}
We train our method on two datasets: FFHQ~\cite{karras2019style} and AFHQv2-CATS~\cite{choi2020stargan}
, which contain 70K human face images of $1024^2$ resolution and 5.5K cat face images of $512^2$ resolution, respectively. Camera poses are estimated using off-the-shelf landmark detectors \cite{bulat2017far,BradCatHipsterize}. 24 surface manifolds are used in GRAM-HD for both human face and cats as in GRAM~\cite{deng2021gram}.
All our models are trained on 8 NVIDIA Tesla V100 GPUs with 16GB memory. More training details such as learning rate and batchsize are presented in the \emph{suppl. mateiral}.

\subsection{Visual Results}
\paragraph{Generated images}
Figure~\ref{fig:results} shows some images samples generated by our method. GRAM-HD can generate high-quality images at high resolution with rich details. Moreover, it allows explicit manipulation of camera pose while maintains strong 3D consistency across different views. For some thin structures like human hair, glasses, and whiskers of cats, our method produces realistic fine details and correct parallax effect viewed from different angles. 

\vspace{-12pt}
\paragraph{Radiance maps and 3D geometry}

Figure \ref{fig:superres} (left) shows the rendered images and radiance maps on manifold surfaces before and after super-resolution. Visually inspected, the HR images and radiance maps are consistent with the LR counterparts in general but contain much more details and thin structures, especially for those high-frequency area like hair and fur.
Figure \ref{fig:superres} (right) presents the 3D proxy shapes extracted for some generated instances.
Here we use a multi-view depth fusion method to extract the shapes since at high resolution the grid sampling and mesh extraction strategy used in \cite{deng2021gram} is not applicable; details can be found in the \emph{suppl. material}. 
As we can see, the shapes extracted at high resolution contain much more shape details.

\begin{figure*}[t!]
	\centering
	\includegraphics[width=1.0\textwidth]{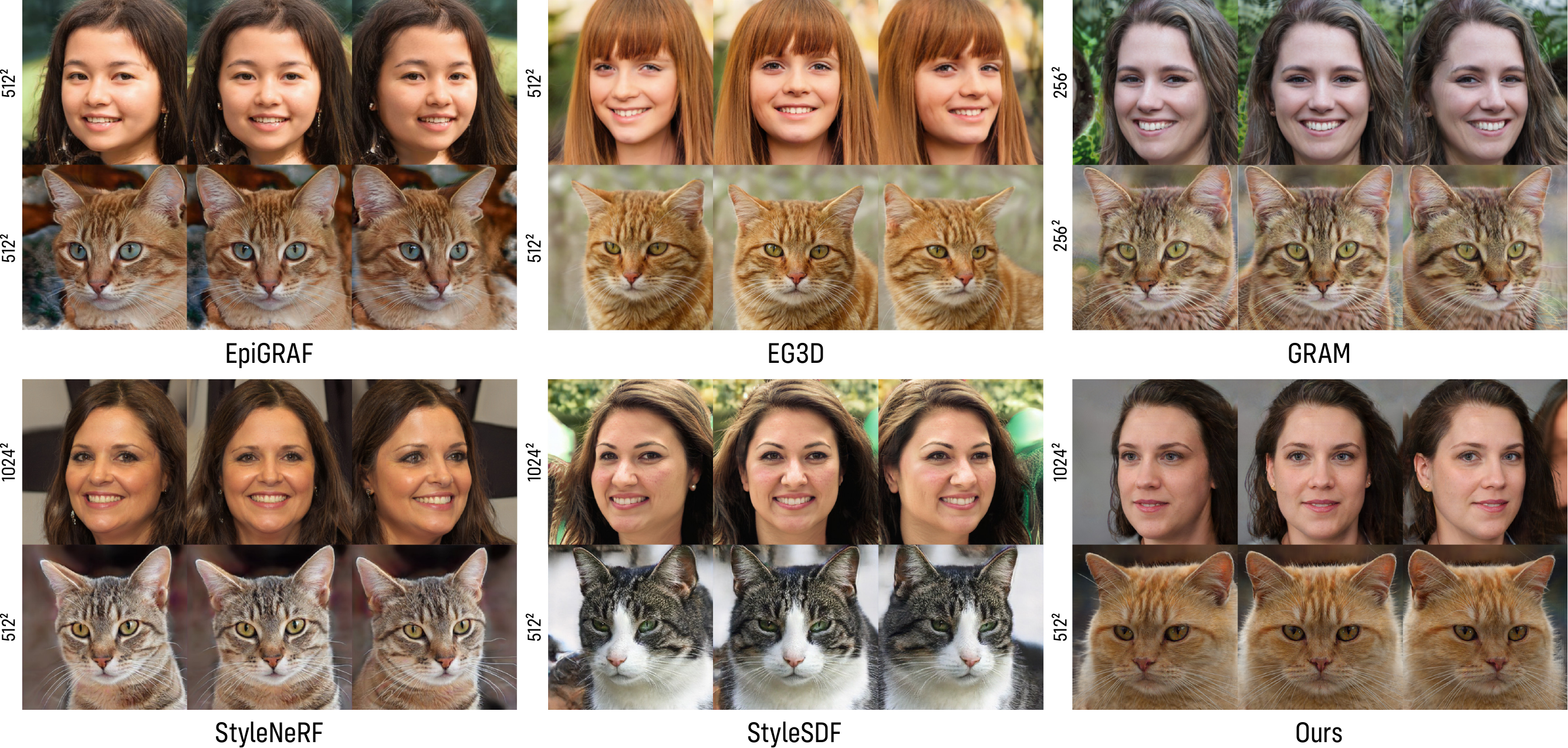}
	\vspace{-20pt}
	\caption{Qualitative comparison with recent 3D-aware GANs. The cat images of StyleNeRF are taken from their paper which is produced by a model trained on all images in AFHQv2; our training of StyleNeRF on cat images failed. (\textbf{Best viewed with zoom-in})}
	\label{fig:comparison}
	\vspace{-4pt}
\end{figure*}
\begin{figure*}[t!]
	\centering
	\includegraphics[width=1.0\textwidth]{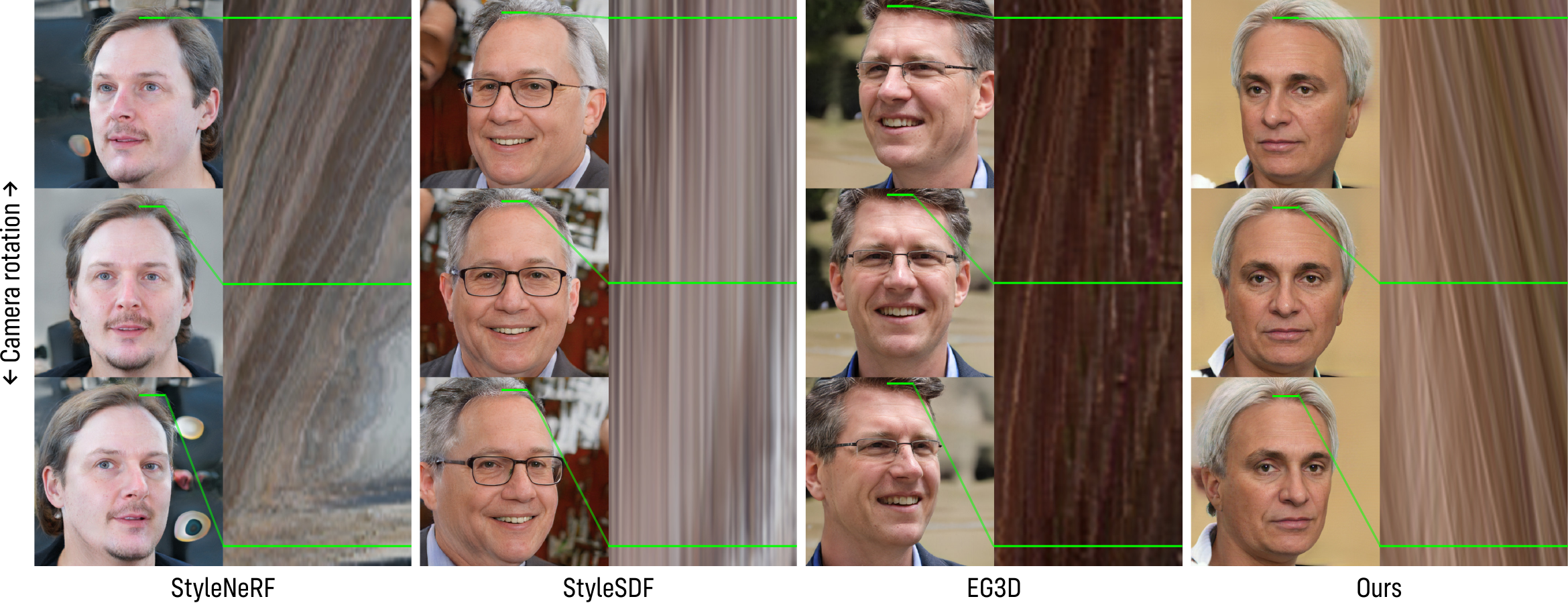}
	\vspace{-20pt}
	\caption{
	Comparison of 3D consistency using spatiotemporal line textures akin to the Epipolar Line Images (EPI)~\cite{bolles1987epipolar}. 
	We rotate the camera horizontally and stack the texture of a fixed horizontal line segment. Our method leads to a natural and smooth texture pattern, whereas others yield distorted and/or noisy patterns, indicating different degrees of 3D inconsistency.  
	Detailed explanations can be found in the text.
	(\textbf{See the accompanying video for results under continuous view change.})}
	\label{fig:comparison_consistency}
	\vspace{-8pt}
\end{figure*}

\subsection{Comparison with Previous Methods}

We compare our GRAM-HD with several recent 3D-aware GANs that can generate high resolution images, including  StyleNeRF~\cite{gu2021stylenerf}, StyleSDF~\cite{or2022stylesdf}, EG3D~\cite{chan2022efficient} and EpiGRAF~\cite{skorokhodov2022epigraf}. For StyleNeRF and StyleSDF, we use their publicly-released models trained on FFHQ for comparison, and train them with the released code on AFHQv2-CATS. However, our training of StyleNeRF on AFHQv2-CATS failed even though multiple trials were made, so no comparison is made for this setting.
For EG3D, we use their released models. For EpiGRAF, we train with official implementation using identical datasets for comparison.

\vspace{-12pt}
\paragraph{Qualitative comparison}
Figure~\ref{fig:comparison} shows some generated images of different methods: StyleNeRF~\cite{gu2021stylenerf}, StyleSDF~\cite{or2022stylesdf}, EG3D~\cite{chan2022efficient}, EpiGRAF~\cite{skorokhodov2022epigraf}, GRAM~\cite{deng2021gram}, and our GRAM-HD. Visually inspected, the images from GRAM-HD have similar quality to other method that adopts 2D CNN for image-space upsampling and superior to EpiGRAF which is though a pure 3D representation, but lack details due to volume rendering.

\setlength{\arrayrulewidth}{0.5mm}
\setlength{\tabcolsep}{3pt}
\renewcommand{\arraystretch}{1.0}
\begin{table}
	\centering
	\caption{Quantitative comparison of image quality using FID and KID ($\times$1000) between 20K generated and real images.}
	\label{tab:comparison_quality}
	\centering
	\small  
	\vspace{-8pt}
	\begin{tabular}{ccccccccc}
		\toprule
		&\multicolumn{2}{c}{\!\!FFHQ1024\!\!}&\multicolumn{2}{c}{\!\!FFHQ512\!\!}&\multicolumn{2}{c}{\!CATS512\!}&\multicolumn{2}{c}{\!FFHQ256\!}\\
		Method     & FID &KID & FID &KID &FID &KID &FID &KID\\
		\midrule
		\!\!StyleNeRF\,\cite{gu2021stylenerf}\!\!   &      9.45      & \textbf{2.65} & -- & -- &      --        &       -- & \textbf{9.24} & \textbf{3.19}       \\
		\!\!StyleSDF\,\cite{or2022stylesdf}\!\!    & \textbf{9.44}  &      2.83     & -- & -- &      7.91        &       3.90    & -- & --    \\
		\!\!EG3D\,\cite{chan2022efficient}\!\!    &  -- & -- & \textbf{8.72}  &     \textbf{3.61}     &     \textbf{6.28}        &       \textbf{1.67}    & -- & --    \\
        \midrule
        \!\!EpiGRAF\,\cite{skorokhodov2022epigraf}\!\! & -- & -- & 14.7 & 6.93 & 9.40 & 4.11 & 16.1 & 7.73 \\
		\emph{Ours}        &      12.0      &     5.23      & 12.2 & 5.41 &     7.67      &     3.41 &  11.8 & 4.72   \\
		\bottomrule
	\end{tabular}
	\vspace{-4pt}
\end{table}

\begin{table}
	\caption{Quantitative comparison of 3D consistency measured by the multiview reconstruction quality of the NeuS~\cite{wang2021neus} method.}
	\label{tab:comparison_consistency}
	\centering
	\small
	\vspace{-8pt}
    \begin{tabular}{ccccccccc}
		\toprule
		&\multicolumn{2}{c}{FFHQ1024}&\multicolumn{2}{c}{FFHQ512}&\multicolumn{2}{c}{CATS512}&\multicolumn{2}{c}{FFHQ256}\\
		Method & \!\!\! \footnotesize PSNR \!\!\! &\!\!\! \footnotesize SSIM \!\!\! & \!\!\! \footnotesize PSNR \!\!\! &\!\!\! \footnotesize SSIM \!\!\! & \!\!\! \footnotesize PSNR \!\!\! &\!\!\! \footnotesize SSIM \!\!\!& \!\!\! \footnotesize PSNR \!\!\! &\!\!\! \footnotesize SSIM \!\!\!\\
		\midrule
		\!\!StyleNeRF\,\cite{gu2021stylenerf}\!\!\!   &      30.0      &     0.80     & -- & -- &     --     &  -- & 31.9 & 0.92 \\
		\!\!StyleSDF\,\cite{or2022stylesdf}\!\!\!    &      31.1      &     0.84     & -- & -- &      26.6       &       0.75    & -- & --  \\
		\!\!EG3D\,\cite{chan2022efficient}\!\!    &  -- & -- & 33.7  &     0.88     &   28.4 &  0.78 & -- & --    \\
		\emph{Ours} & \textbf{33.8}  & \textbf{0.87}& \textbf{34.0} & \textbf{0.90} &  \textbf{28.8} &  \textbf{0.81} & \textbf{36.5} &  \textbf{0.96} \\
		\bottomrule
	\end{tabular}
	\vspace{-4pt}
\end{table}

\setlength{\arrayrulewidth}{0.5mm}
\setlength{\tabcolsep}{3pt}
\renewcommand{\arraystretch}{1.0}
\begin{table}
	\caption{Quantitative comparison of image quality between GRAM and our GRAM-HD. Due to its high memory cost, GRAM can only train on up to $256^2$ resolution.
 }
	\label{tab:comparison_gram}
	\centering
	\small  
	\vspace{-8pt}
	\begin{tabular}{ccccccc}
	\toprule
	&\multicolumn{2}{c}{FFHQ256}&\multicolumn{2}{c}{CATS256}\\
	Method     & FID&KID&FID&KID\\
	\midrule
	GRAM~\cite{deng2021gram}        &      15.0      &      6.55     &     12.9      &       7.37    \\
	\emph{Ours}         &      \textbf{11.8} & \textbf{4.72} & \textbf{7.05} & \textbf{2.53} \\
	\bottomrule
    \end{tabular}
    \vspace{-8pt}
\end{table}

However, the results of other methods except for EpiGRAF exhibit different levels of 3D inconsistency under view change, as can be observed in our accompanying video. 
To better visualize 3D consistency here, we present the spatiotemporal textures of different methods in Figure~\ref{fig:comparison_consistency}. Specifically, we smoothly
rotate the camera horizontally and stack the texture of a fixed horizontal line segment, forming some spatiotemporal texture images similar to the Epipolar Line Images (EPI)~\cite{bolles1987epipolar}. For 3D-consistent generation, the resultant spatiotemporal texture should appear smooth and natural.
Figure~\ref{fig:comparison_consistency} shows that the resultant texture image of StyleNeRF contain both distorted and noisy regions, indicating both low-frequency inconsistency and high-frequency texture flicking. The line texture from StyleSDF barely changes across views, indicating the texture sticking  artifact~\cite{karras2021alias}. EG3D's spatiotemporal texture also contains some non-smooth, high-frequency patterns which are caused by texture flicking. In contrast, the texture from GRAM-HD appears natural with no noticeable noise or distortion, demonstrating its strong 3D consistency. 

\vspace{-12pt}
\paragraph{Quantitative comparison}
Table~\ref{tab:comparison_quality} shows the FID~\cite{heusel2017gans} and KID~\cite{binkowski2018demystifying} metrics of different methods. 
Our scores are slightly higher than those 3D-inconsistent methods, which indicate again the comparable image generation quality. We further compare GRAM-HD to 3D-consistent methods EpiGRAF and GRAM in Table~\ref{tab:comparison_quality} and Table~\ref{tab:comparison_gram}, respectively. GRAM-HD consistently outperforms on all experiment settings and can be applied to higher resolution. To evaluate computation efficiency, we record the running time and memory cost for each method in Table~\ref{tab:comparison_compute}, where GRAM-HD is at most 3 times slower than 3D-inconsistent methods but significantly faster than 3D-consistent methods, especially at high resolution. This demonstrates that GRAM-HD not only has superior computation efficiency but also generates higher-quality images. 

To quantitatively evaluate 3D consistency, for each method we generate 30 images under different views, and train the multiview reconstruction method NeuS~\cite{wang2021neus} on them. We report the PSNR and SSIM scores of the images reconstructed by NeuS. In theory, the more consistent the input mutlview images are, the higher the reconstruction quality will be. Table~\ref{tab:comparison_consistency}  presents the scores averaged on 50 randomly generated instances. As expected, our GRAM-HD leads to consistently better reconstruction than StyleNeRF, StyleSDF and EG3D, demonstrating its superior multiview consistency compared to the competing methods. 

\setlength{\arrayrulewidth}{0.5mm}
\setlength{\tabcolsep}{3pt}
\renewcommand{\arraystretch}{1.0}
\begin{table}[t!]
	\small
  \caption{Comparison of running time (\emph{left}) and memory cost (\emph{right}) on a NVIDIA V100 GPU. Whole computational graph retained.
  	``Strict 3D" means methods that do not use 2D upsampling thus have strong mutliview consistency.}
  \label{tab:comparison_compute}
  \vspace{-6pt}
  \centering
    \begin{minipage}{.28\textwidth}
  	\centering
  	\begin{tabular}{cccccc}
  		\toprule
  		& &\!Methods\!                 & $256^2$ & $512^2$ & \!$1024^2$\! \\
  		\midrule
  		\multicolumn{2}{c}{\multirow{3}{*}{\rotatebox{90}{\footnotesize Strict 3D}}}~  &piGAN & 0.29s &  1.15s &  4.59s   \\
  		& &GRAM&  0.43s &  1.54s & 6.69s    \\
  		& &\emph{Ours}  &   0.18s  & 0.22s &   0.36s  \\
  		\midrule
  		\!\multirow{3}{*}{\rotatebox{90}{\footnotesize Not }}\!\!&\!\multirow{3}{*}{\rotatebox{90}{\footnotesize strict 3D}}~~&\!\!StyleNeRF\!\! & 0.06s&	0.08s&	0.16s\\
  		& &StyleSDF  & 0.11s & 0.11s	&	0.13s  \\
  		& &EG3D   &--	&0.10s&	-- \\
  		
  		\bottomrule
  	\end{tabular}
  \end{minipage}
  \begin{minipage}{.18\textwidth}
  	\centering
  	\begin{tabular}{ccc}
  		\toprule
  		$256^2$ & $512^2$ & \!$1024^2$\! \\
  		\midrule
  		21.1G & OOM & OOM\\
  		22.2G & OOM	& OOM\\
  		 5.4G &	6.3G & 9.0G	\\
  		\midrule
  		0.48G  & 0.85G  & 3.0G \\
  		0.40G&	0.53G&	0.80G\\
  		--& 2.96G & --   \\
  		\bottomrule
  	\end{tabular}
  \end{minipage}
  \vspace{-4pt}
\end{table}

\begin{table}
	\caption{Ablation study on FFHQ256.}
	\label{tab:ablation}
	\centering
	\small
	\vspace{-8pt}
    \begin{tabular}{lc}
	\toprule
	Method        & FID\\
	\midrule
	StyleGAN2 architecture & 18.2\\
	Base architecture & 14.4\\
	\ + Sub-pixel convolutions & 14.1\\
	\ + Style modulation & 14.0\\
	\ + LR generator feature & 13.3\\
	\ + Background-net & 12.0\\
	\ + $\mathcal{L}_\mathrm{patch}$ (\emph{Ours}) & \textbf{11.8}\\
	\ - $\mathcal{L}_\mathrm{cons}$ & 16.9\\
	\bottomrule
    \end{tabular}
    \vspace{-8pt}
\end{table}

\begin{figure*}[t!]
	\centering
	\includegraphics[width=1\textwidth]{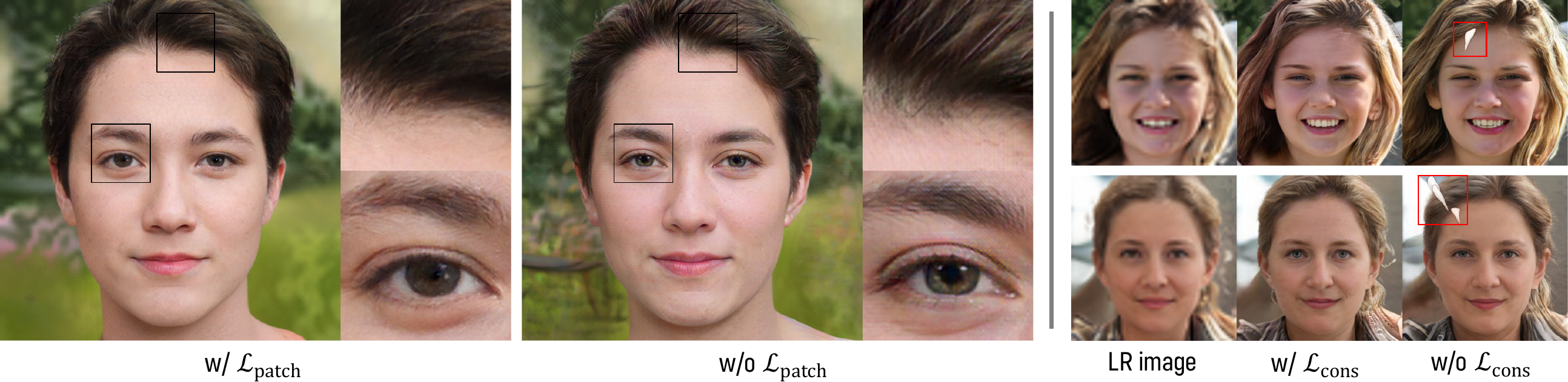}
	\vspace{-20pt}
	\caption{\textbf{Left:} Sample results w/ and w/o the patch adversarial loss $\mathcal{L}_\mathrm{patch}$ on $1024^2$ resolution. $\mathcal{L}_\mathrm{patch}$ can effectively eliminate checkerboard artifacts.
		\textbf{Right:} Sample results w/ and w/o cross-resolution consistency loss $\mathcal{L}_\mathrm{cons}$. Some unwanted floaters appear in front of the faces without $\mathcal{L}_\mathrm{cons}$.}
	\label{fig:ablation}
	\vspace{-4pt}
\end{figure*}
\begin{figure*}[t!]
	\centering
	\includegraphics[width=1\textwidth]{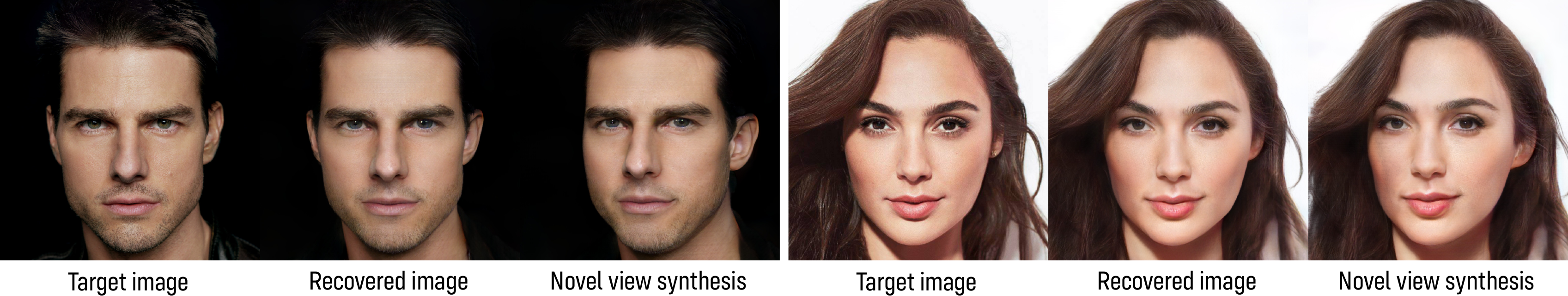}
	\vspace{-20pt}
	\caption{High-resolution ($1024^2$) image embedding and editing results (\textbf{Best viewed with zoom-in})}
	\label{fig:image_embedding}
	\vspace{-8pt}
\end{figure*}

\subsection{Ablation Study}

We further conduct ablation studies to validate the efficacy of our super-resolution CNN architecture design and the loss functions for high-resolution training. Unless otherwise specified, the experiments are conducted on FFHQ with $256^2$ resolution for efficiency.

\vspace{-12pt}
\paragraph{Network architecture}
The first 5 rows of Table~\ref{tab:ablation} shows the FID scores with different architectures we tested for the super-resolution CNN.  
Here the base architecture is the original network of ESRGAN~\cite{wang2018esrgan}, which we found to perform better than a decoder structure of StyleGAN2~\cite{karras2020analyzing}. We then add different components, including subpixel convolutions for upsampling, style modulation, intermediate feature from LR radiance generator, and a separate CNN for background radiance, which all lead to lower FID scores.

\vspace{-12pt}
\paragraph{Loss functions}
We verify the efficacy of the two new loss functions $\mathcal{L}_\mathrm{patch}$ and $\mathcal{L}_\mathrm{cons}$ with both numerical metrics and visual quality. The results are presented in Table~\ref{tab:ablation} (last two rows) and Figure~\ref{fig:ablation}, respectively. We find that adding the patch discriminator and applying $\mathcal{L}_\mathrm{patch}$ lead to lower FID score at low resolution ($256^2$ in Table~\ref{tab:ablation}), and can effectively eliminate the checkerboard artifact at high resolution ($1024^2$ in Figure~\ref{fig:ablation}). Without the cross-resolution consistency loss $\mathcal{L}_\mathrm{cons}$, the FID score increases significantly and the super-resolution CNN generates some floaters in the front as shown in Figure ~\ref{fig:ablation}, which might be caused by less stable training.

\subsection{Applications}

\paragraph{High resolution image embedding and editing} Like other GAN methods, one can embed an image into the latent space of our trained GRAM-HD and achieve pose editing by rendering images at novel views. As shown in Figure~\ref{fig:image_embedding}, our method can faithfully reconstruct a high-resolution image through GAN inversion and generate high-fidelity novel view rendering results.

\vspace{-12pt}
\paragraph{Real-time free-view synthesis}
Similar to GRAM~\cite{deng2021gram}, we can cache the manifold surfaces and HR radiance maps as textured 3D meshes and then run fast free-view synthesis with mesh rendering. 
With an efficient mesh rasterizer from \cite{laine2020modular}, we achieve free-view synthesis of $1024^2$ images at 90 FPS on a Nvidia Tesla V100 GPU.

\section{Conclusion}\label{sec:conclusion}
We have presented a novel GAN approach for high-resolution, 3D-consistent multiview image generation, which is trained on unstructured single image collections. Our key idea is to tackle a 3D super-resolution task using efficient 2D CNNs by leveraging the recent radiance manifold representation. The experiments show that our generation results at high resolution (\eg, $1024^2$) are not only of high quality but also strongly 3D consistent, significantly outperforming recent 3D-ware GANs. We believe our method bring 3D-consistent image generation closer to the traditional 2D GANs, and it paves the way for high-quality 3D content creation applications such as 3D video generation and animation. 

\vspace{-12pt}
\paragraph{Limitations and future work}

Our method still have several limitations. Based on the radiance manifold representation, it is difficult to handle objects with complex 3D geometry. Its view extrapolation capability is also limited compared to dense 3D radiance fields. Besides, the generation quality of our method still lags behind traditional 2D image generative models. Better representations or training strategies could be further explored to close the gap.

\vspace{-12pt}
\paragraph{Ethics consideration} The goal of this paper is to generate images of virtual objects for applications such as photorealisitc virtual avatar creation. However, it could be misused to create misleading or harmful contents.
We condemn any behavior to use our method for fraud purposes. On the positive side, our method could be used for forgery detection system testing and for generating privacy-free contents. The face dataset we use may contain certain biases in race, gender, etc., which can be inherited by our trained model and results. Bias-free generative modeling is important direction worth further exploration by the community.

{\small
\bibliographystyle{ieee_fullname}
\bibliography{egbib}
}

\clearpage

\appendix

\begin{strip}
\centering
\Large{\textbf{Supplementary Material}}
\end{strip}

\renewcommand{\thesection}{\Alph{section}}
\renewcommand{\thefigure}{\Roman{figure}}
\renewcommand{\thetable}{\Roman{table}}
\renewcommand{\theequation}{\Roman{equation}}
\setcounter{figure}{0}
\setcounter{equation}{0}

\begin{figure*}[h]
\vspace{20pt}
	\centering
	\includegraphics[width=0.8\textwidth]{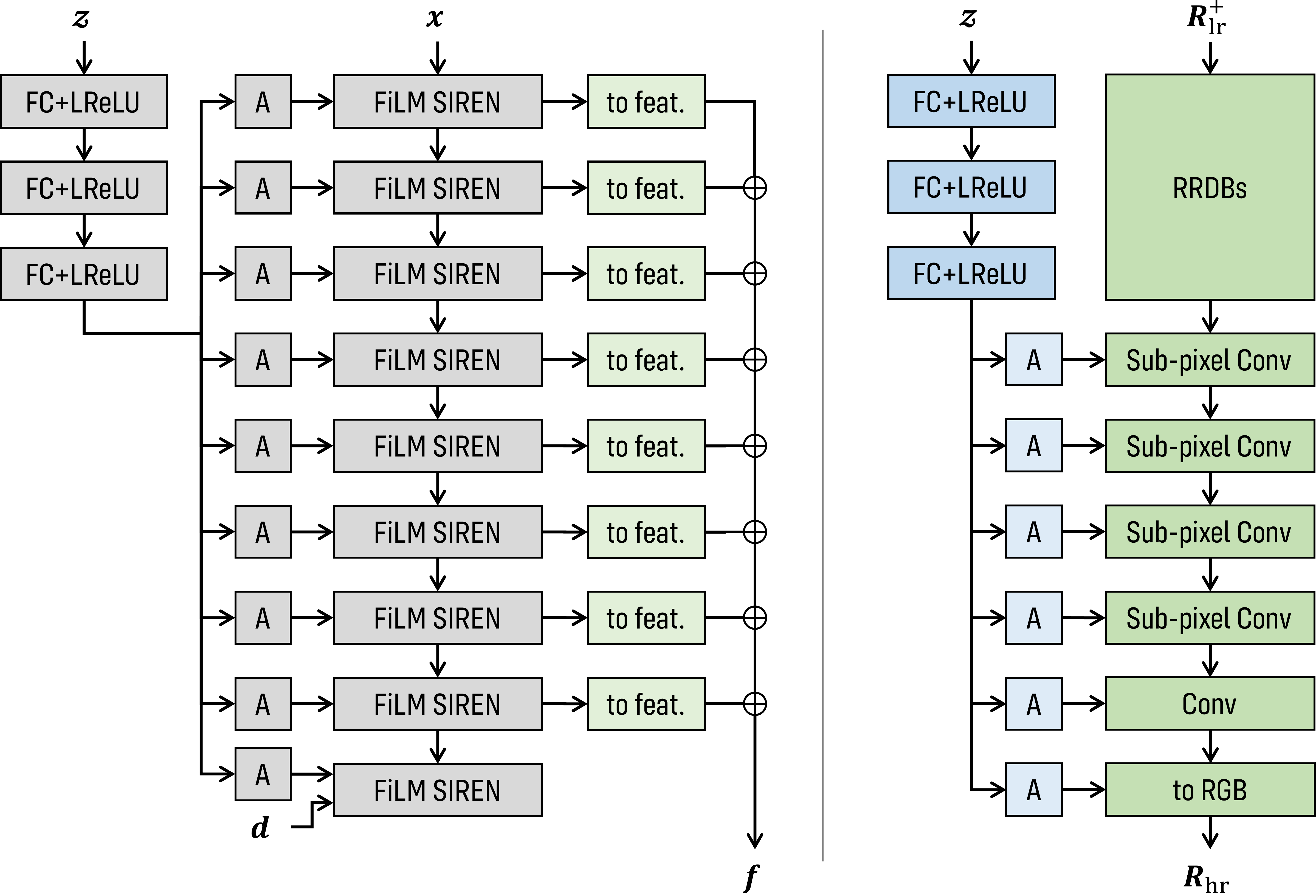}
	\caption{Network architecture. The components in gray color are from GRAM~\cite{deng2021gram} and others are newly introduced in our GRAM-HD.  \textbf{Left:} The architecture of the radiance manifold generator. The intermediate features are transformed by simply fully connected layers and accumulated to be the input for the super-resolution CNN. The RGB$\alpha$ prediction branches and the manifold predictor network are the identical to GRAM and omitted here for brevity. \textbf{Right:} The architecture of the super-resolution module with several Residual-in-Residual Dense Blocks (RRDBs)~\cite{wang2018esrgan} and sub-pixel convolutional layers~\cite{shi2016real} as backbone.}
	\label{fig:architecture}
\end{figure*}

\section{Network architecture}
\paragraph{Radiance manifold generator}
As shown in Figure~\ref{fig:architecture} (left), the network architecture of our radiance manifold generator is the same as GRAM~\cite{deng2021gram}, except that we additional apply several fully-connected layers with skip connections to extract intermediate features. These layers project the 256-dimension hidden features of the FiLM SIREN MLP~\cite{chan2021pi} into 32-dimension intermediate features which are used as the input for the super-resolution module.

\paragraph{Super-resolution module}
Figure \ref{fig:architecture} (right) shows the detailed structure of our super-resolution module. For LR feature processing, we apply 8 RRDB blocks, each having 64 channels. We then apply 4 sub-pixel convolution layers with 64, 64, 32, and 16 channels respectively for upsampling to $1024^2$ resolution. 
Finally, a 16-channel convolution layer and a 4-channel projection layer are applied to produce the color and occupancy. Besides, a mapping MLP with three 256-dimension hidden layers maps the latent code to the style code. Each conv layer after the RRDBs is modulated by an affine-transformed style code. 

\section{More implementation details}
\subsection{Data preparation}
We align the images in FFHQ~\cite{karras2019style} and AFHQv2-CATS~\cite{choi2020stargan} using detected landmarks. Specifically, we first detect landmarks of the images (5 landmarks for FFHQ and 9 for AFHQv2-CATS) using of-the-shelf landmark detectors~\cite{bulat2017far,BradCatHipsterize}. We then resize and crop the images by solving a least-square fitting problem between the detected landmarks and a set of predefined 3D keypoints, following the strategy of~\cite{deng2019accurate}. The 3D keypoints for human face are derived from the mean face of a 3D parametric model~\cite{paysan20093d}, while for cat they are some manually-selected vertices on a cat head mesh downloaded from the Internet.

We extract camera poses for the images in the datasets, which are used to estimate the prior camera pose distributions and serve as the pseudo labels for the pose loss term following \cite{deng2021gram}. For FFHQ, the face reconstruction method of~\cite{deng2019accurate} is employed for pose estimation. For AFHQv2-CATS, the estimated angles are obtained via solving the aforementioned least-square fitting. Then, we fit Gaussian distributions on yaw and pitch angles as prior pose distributions. The standard deviations for yaw and pitch angles are $(0.3, 0.15)$ and $(0.18, 0.15)$ for FFHQ and AFHQv2-CATS, respectively.

\subsection{More training details}
During training, we randomly sample latent code $\boldsymbol z$ from the normal distribution and camera pose $\boldsymbol\theta$ from the estimated prior distributions of the datasets.
A two-stage training strategy is applied as mentioned in the main paper.

At the first training stage, we initialize the radiance manifold generator following \cite{deng2021gram}: 23 evenly distributed sphere manifolds centered at $(0,0,-1.5)$ covering 3D objects inside of the $[-1,1]^3$ cube and an additional background plane at $z=-1$ are initialized. The learning rates are set to $2\mathrm{e}-5$ for the radiance manifold generator and $2\mathrm{e}-4$ for the LR discriminator.

At the second stage, we freeze the radiance manifold generator and only optimize the super-resolution module (and the newly-added  transformation layers for intermediate features). The HR discriminator is trained from scratch without progressive growing.
The learning rates are set to $2\mathrm{e}-4$ for both the super-resolution module and the HR discriminator.

For all the training processes, we use the Adam~\cite{kingma2015adam} optimizer with $\beta_1=0$ and $\beta_2=0.9$. Batch size is set to 32 in the first training stage and 32, 16, 16 in second stage for $256^2$, $512^2$, $1024^2$, respectively. 
We train our model for 100K iterations on FFHQ and 40K iterations on AFHQv2-CATS since it is a relatively small dataset. Training took 3 to 7 days depending on the dataset and resolution.

\paragraph{Background super-resolution details}
For the last surface manifold, \emph{i.e.,} the background plane, we use a larger projection view during manifold gridding, as it spans a much wider region for covering the background under extreme views.
A nonlinear mapping is applied to do the sampling. 
Specifically, we first uniformly sample $xy$ coordinates within $[-1,1]^2$ 
and then apply the following nonlinear mapping function:
\begin{equation}
\mathrm{BGTrans}(x)=\left\{
\begin{aligned}
&2\tan(x+0.5)-1\quad&x<-0.5\\
&2x&-0.5\le x\le 0.5\\
&2\tan(x-0.5)+1&x>0.5
\end{aligned}.
\right.
\end{equation}
The purpose of this transformation is to enlarge the sampling region and sample denser points around center. Radiance are calculated at these sampled and transformed coordinates and form the radiance map for super-resolution.
A small independent super-resolution CNN is applied since the radiance distribution on the background significantly differs from the foreground.

\subsection{3D-consistency metric details}
To quantitatively evaluate 3D consistency, we use the reconstruction quality of a recent surface-based multiview reconstruction method - NeuS~\cite{wang2021neus}. Specifically, for each method, we first randomly generate 50 instances. For each instance, we render 30 images with yaw angle evenly sampled from $-0.4$ radian to $0.4$ radian, and train a NeuS model with these images as input. The mean PSNR and SSIM scores of the reconstructed images by NeuS are used as the quantitative metrics. In theory, the more consistent the input mutlview images are, the higher the reconstruction quality will be. For NeuS training, we use the official implementation with default settings.

\subsection{Shape extraction details}
We employ a multiview depth fusion method to extract shapes at high resolution. For a given view, the depth map can be calculated by:
\begin{equation}
\begin{split}
    d({\boldsymbol r}) &= \sum_{i=1}^{N}T({\boldsymbol x}_i)\alpha({\boldsymbol x}_i) z(\boldsymbol x_i) \\
    &= \sum_{i=1}^{N}\prod_{j<i}(1-\alpha({\boldsymbol x}_j))\alpha({\boldsymbol x}_i)z(\boldsymbol x_i),
\end{split}
\end{equation}
where $\boldsymbol r$ is a viewing ray, $\boldsymbol x_i$ are the point samples, \emph{i.e.}, ray-manifold intersections, and $z(\cdot)$ denotes the projected depth.
We then calculate a discrete occupancy field on a 3D sampling grid. Specifically, for each point $\boldsymbol x_\mathrm{s}$ on the sampling grid, we project it to the depth map and calculate its occupancy as $\alpha=\mathrm{Sigmoid}\big(k\big(z(\boldsymbol x_\mathrm{s}) - d\big)\big)$ where $k$ is a scaling factor we set to $10$. We average the occupancy from 15 different views, and run MarchingCube~\cite{lorensen1987marching} to extract the shape.

\subsection{Image embedding details}
Given a target image $I_{\mathrm{t}}$, we freeze the weights of the generator and optimize the style code $w_i$ for each modulated layer to generate an image $I_\mathrm{g}$ that best matches the target image. The following objective function is used:
\begin{equation}
\begin{split}
    \mathcal L_{\mathrm{emb}}&=\|I_\mathrm{g}-I_\mathrm{t}\|^2+ \mathrm{LPIPS}(I_\mathrm{g},I_\mathrm{t})\\
    &+(1-\langle f_\mathrm{id}(I_\mathrm{g}), f_\mathrm{id}(I_\mathrm{t}) \rangle)\\
    &+\sum_i\|w_i-\bar{w}\|^2+\sum_{j}\|\sigma_j^{\mathrm{d}}\|^2,
\end{split}
\end{equation}
where $f_\mathrm{id}$ is an identity feature extractor~\cite{deng2019arcface}, $\mathrm{LPIPS}$ is a perceptual loss from \cite{zhang2018unreasonable}, $\bar{w}$ is the precomputed mean style code,  and $\sigma_j^{\mathrm{d}} =  \mathrm{sqrt}\big(\sum_{i=1}^{N}T({\boldsymbol x}_i)\alpha({\boldsymbol x}_i) z^2(\boldsymbol x_i)-d^2(\boldsymbol r_j)\big)$ is the standard deviation of depth along each viewing ray $\boldsymbol r_j$. The style and depth regularizations are added to avoid overfitting.
With the Adam~\cite{kingma2015adam} optimizer, we first run the optimization on low resolution for 200 steps and then switch to the high resolution for another 5000 steps.

\section{More experimental results}
\subsection{More Qualitative results}

\begin{figure*}
	\centering
	\includegraphics[width=0.83\textwidth]{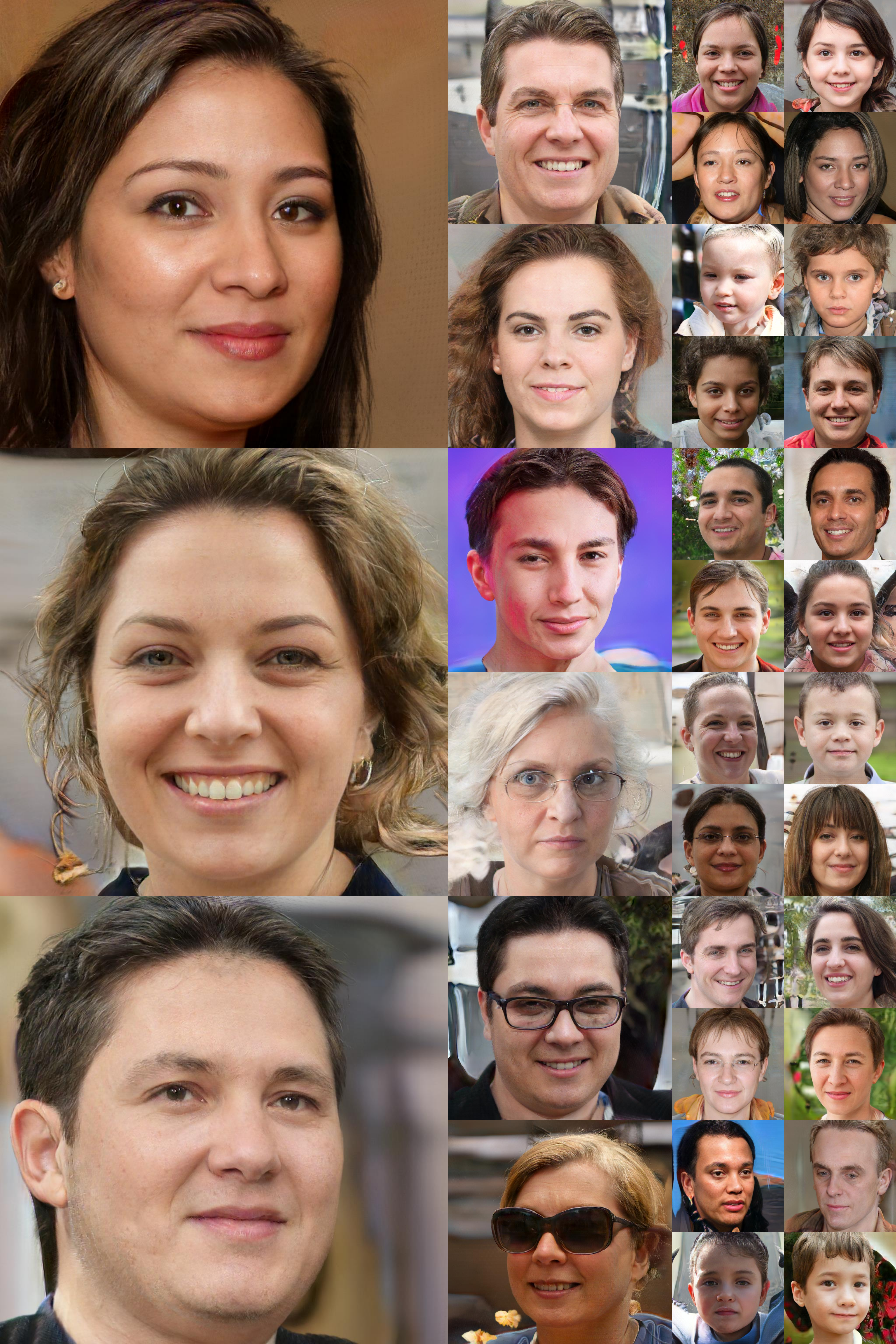}
	\caption{Uncurated $1024^2$ results of GRAM-HD on FFHQ.}
	\label{fig:FFHQ_uncurated}
\end{figure*}

\begin{figure*}
	\centering
	\includegraphics[width=0.83\textwidth]{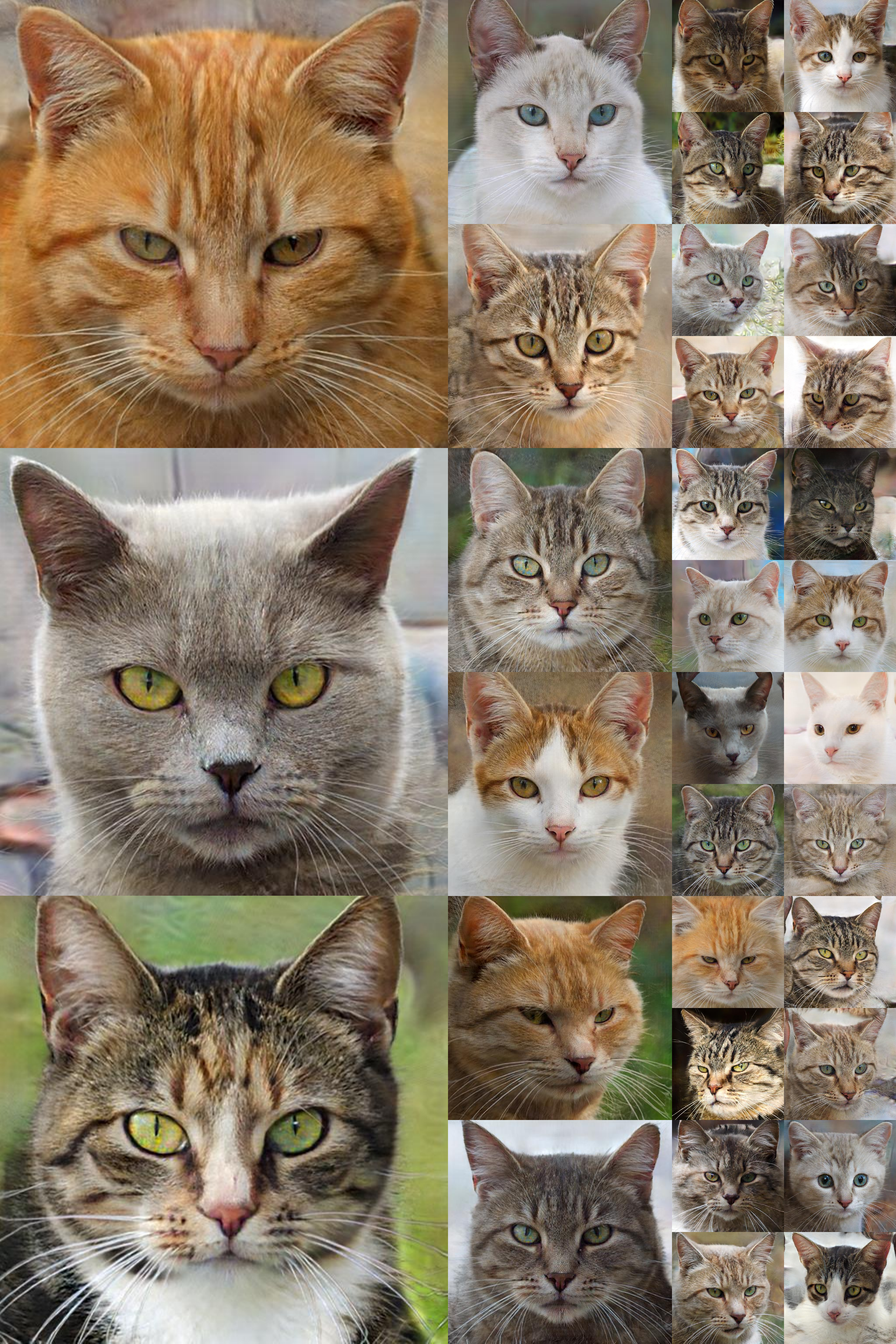}
	\caption{Uncurated $512^2$ results of GRAM-HD on AFHQv2-CATS.}
	\label{fig:AFHQ_uncurated}
\end{figure*}

\begin{figure*}
	\centering
	\includegraphics[width=0.9\textwidth]{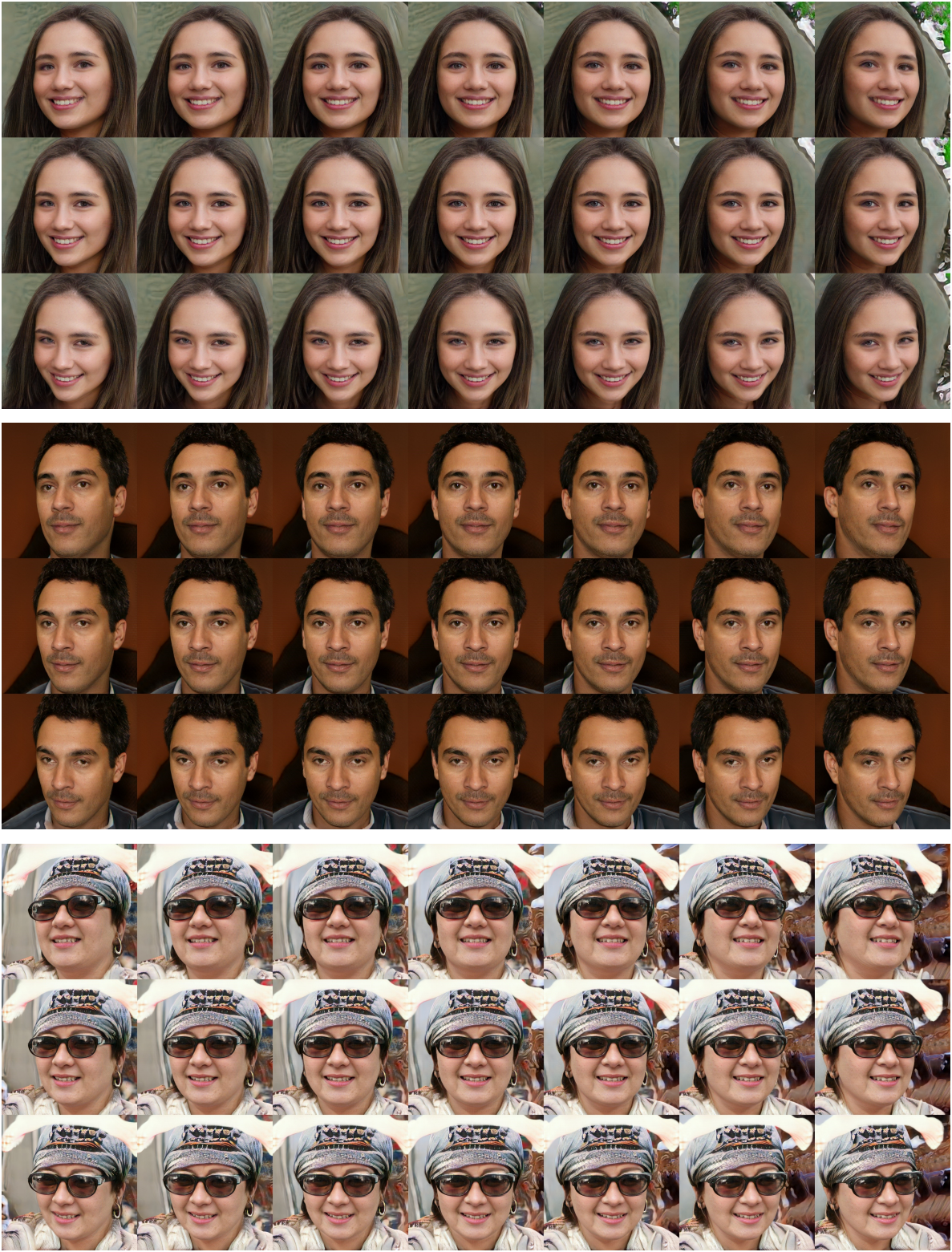}
	\caption{Multiview generation results of GRAM-HD on FFHQ.}
	\label{fig:FFHQ_multiview}
\end{figure*}

\begin{figure*}
	\centering
	\includegraphics[width=0.9\textwidth]{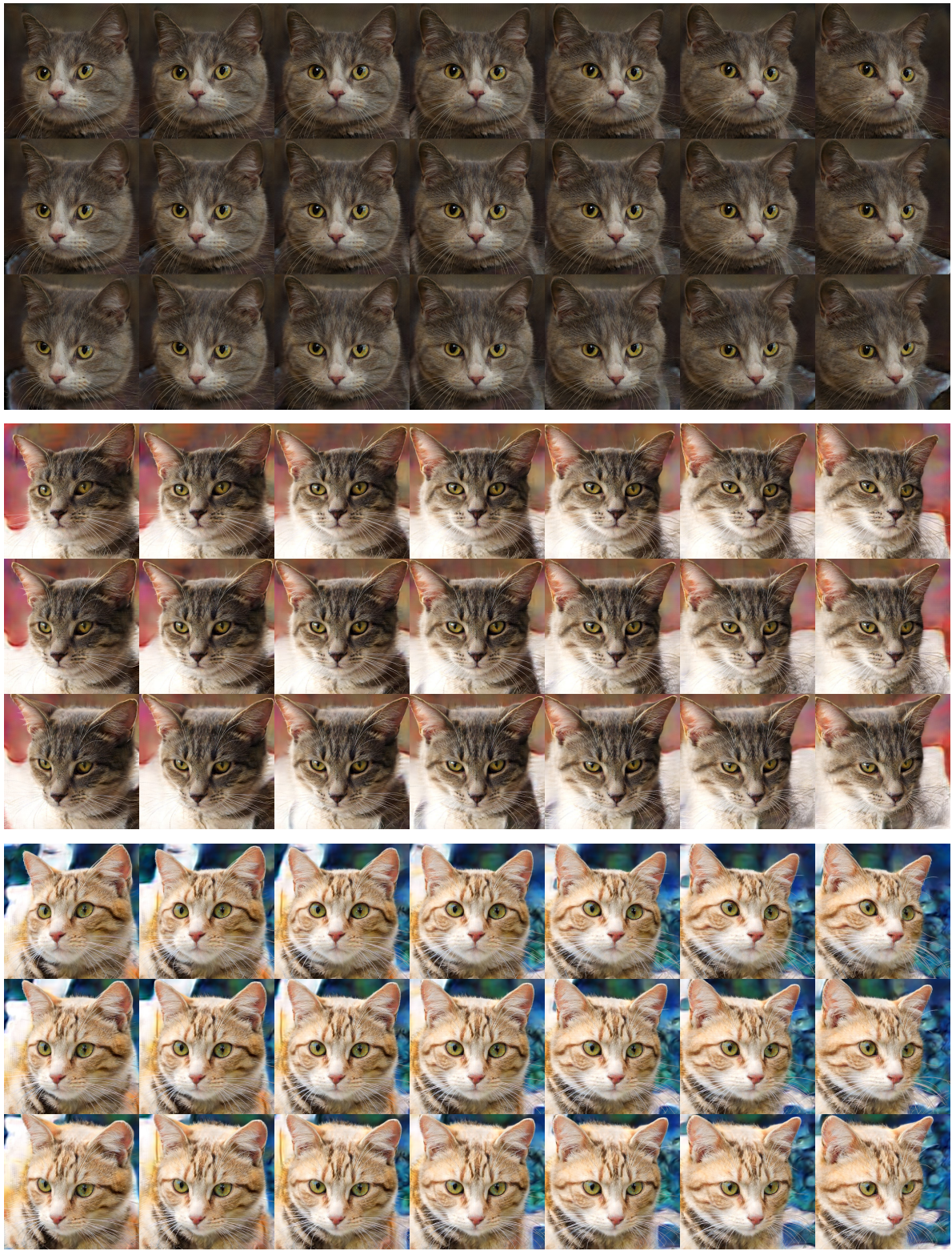}
	\caption{Multiview generation results of GRAM-HD on AFHQv2-CATS.}
	\label{fig:AFHQ_multiview}
\end{figure*}

Figure \ref{fig:FFHQ_uncurated} and \ref{fig:AFHQ_uncurated} present the uncurated generation results of GRAM-HD. Figure \ref{fig:FFHQ_multiview} and \ref{fig:AFHQ_multiview} further show the multiview images of some generated instances. Our method can generate realistic images at high resolution with strong 3D consistency.

\subsection{More Comparisons}
In this section, we provide more comparisons of geometry details and 3D consistency between our method and StyleNeRF~\cite{gu2021stylenerf}, StyleSDF~\cite{or2022stylesdf}, EG3D~\cite{chan2022efficient}, EpiGRAF~\cite{skorokhodov2022epigraf} and GMPI~\cite{zhao2022gmpi}.

\begin{figure*}
	\centering
	\includegraphics[width=1\textwidth]{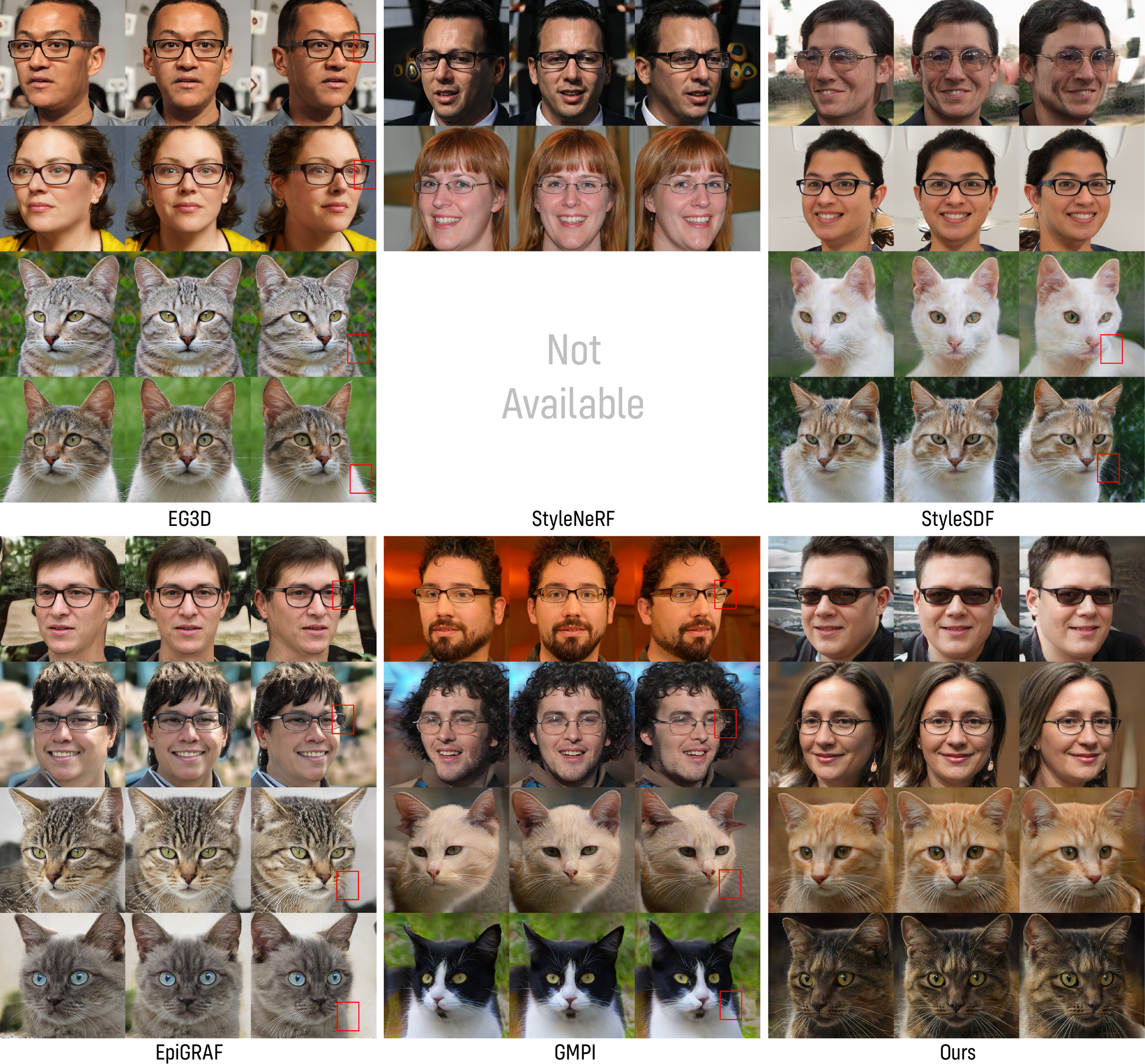}
	\caption{More comparison of geometry details. Our method can generate thin geometry structures such as glasses and whiskers while other methods either suffer some distortion (marked with box) or are not 3D-consistent. (\emph{Best viewed with zoom-in; see also the accompanying video for better visualization.})}
	\label{fig:more_comparison}
\end{figure*}

\paragraph{Visual comparison of geometry details}
In Figure~\ref{fig:more_comparison}, we show more samples from different methods with some thin geometry structure highlighted.
As we can see from the figures, almost all other methods generate some artifacts around eyeglass for human face and whiskers for cats. In particular, all these method failed to generate reasonable whiskers of cats: the generated cat whiskers are stuck onto the cat faces instead of floating naturally in the front. 
In contrast, our method can generate highly-realistic results for such thin structures.

\begin{figure*}
	\centering
	\includegraphics[width=1\textwidth]{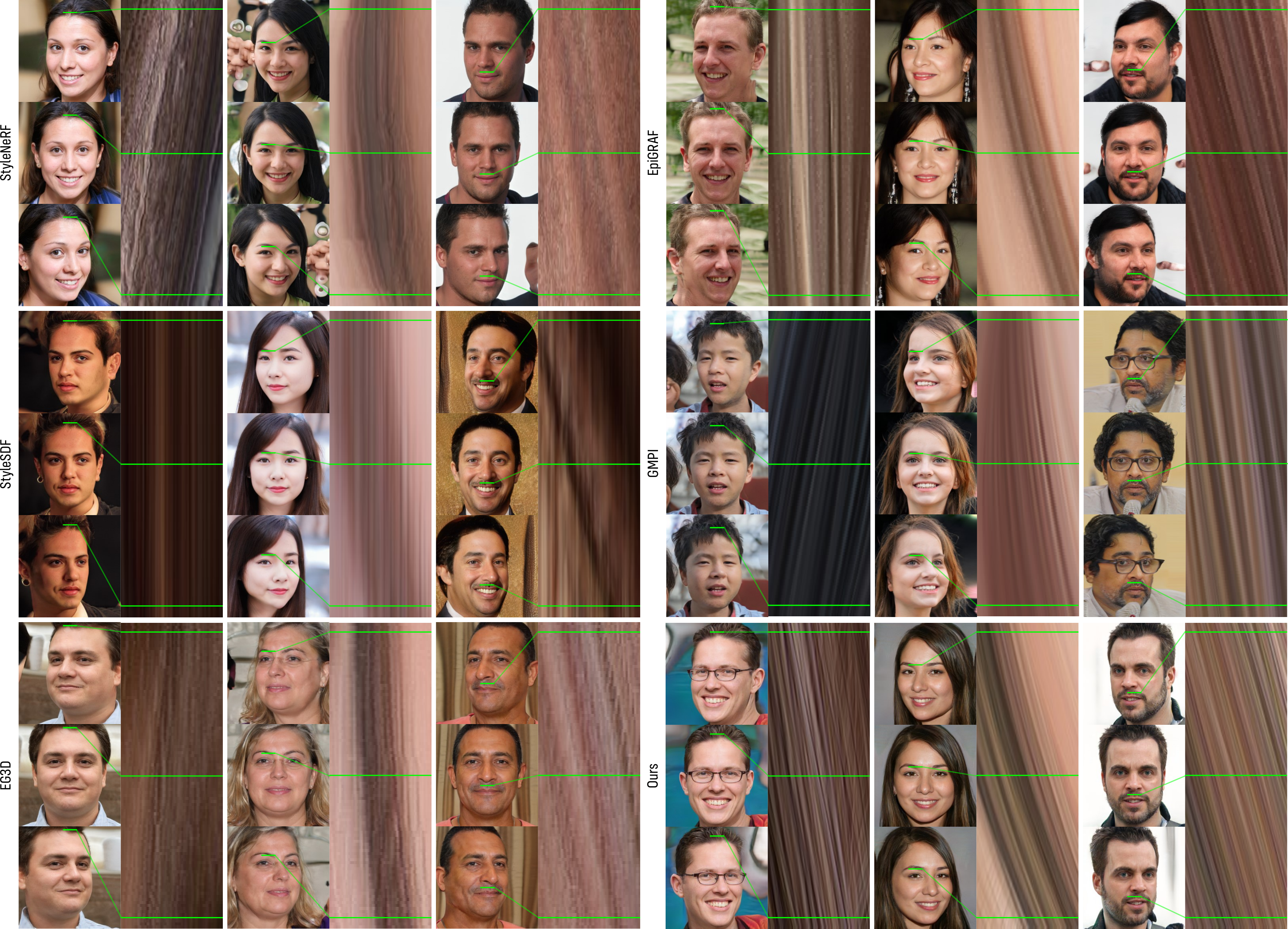}
	\caption{More comparison of 3D consistency using spatiotemporal texture image.}
	\label{fig:FFHQ_epi}
\end{figure*}

\vspace{-8pt}
\paragraph{EPI comparison}
In Figure~\ref{fig:FFHQ_epi}, we show more EPI-like texture images to demonstrate GRAM-HD's superiority on 3D consistency. The textures of StyleNeRF, StyleSDF and EG3D are either distorted or stuck to image coordinates, indicating different types of inconsistency. Although EpiGRAF uses a pure 3D representation without image-space upsampler, there are still some noise on its generated textures as the Monte-Carlo volume rendering is not noise-free. The textures from our method and GMPI are smooth and natural, demonstrating their superior 3D consistency.

\subsection{Latent space interpolation}
\begin{figure*}
	\centering
	\includegraphics[width=1\textwidth]{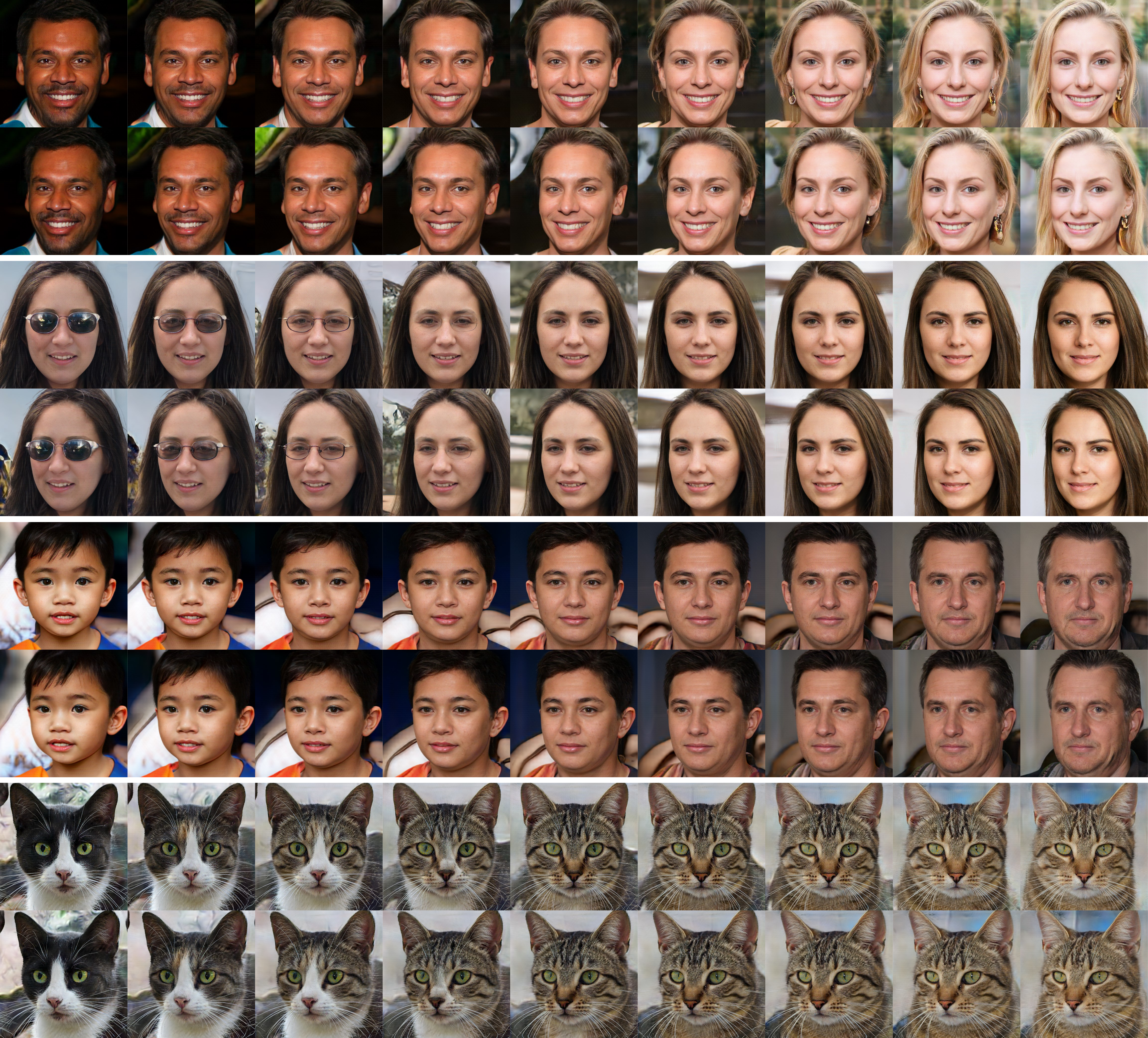}
	\caption{Latent space interpolation results.}
	\label{fig:interp}
\end{figure*}

Figure \ref{fig:interp} shows the results of latent space interpolation with GRAM-HD. We select generated instances of different gender, skin color, age, \emph{etc.}, and then show the results by linearly interpolating their latent codes. The meaningful intermediate results and smooth changes demonstrate the reasonable latent space learned by GRAM-HD.

\subsection{Style mixing}
\begin{figure*}
	\centering
	\includegraphics[width=0.95\textwidth]{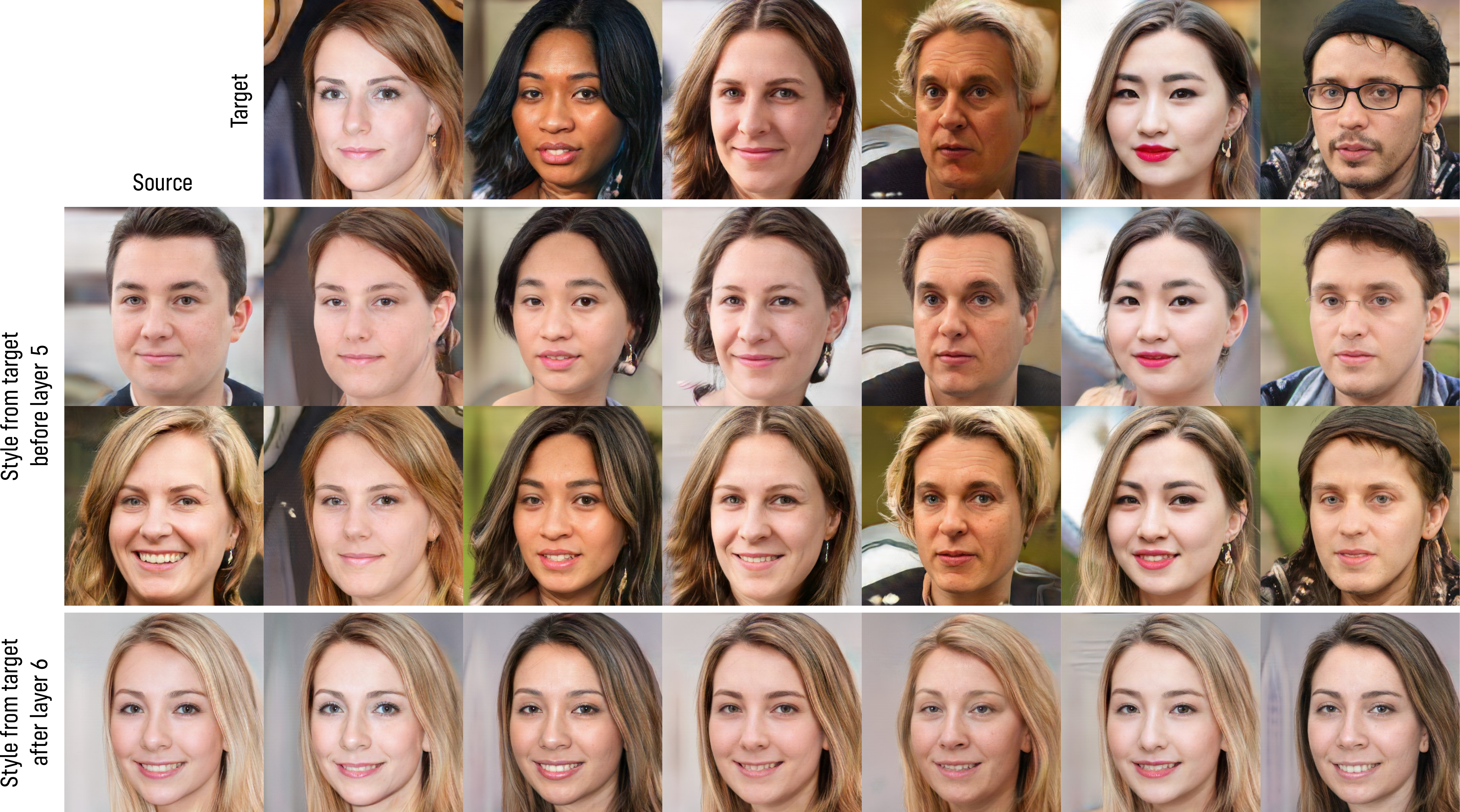}
	\caption{Style mixing between different generated subjects.}
	\label{fig:mixing}
\end{figure*}

We further tested style mixing~\cite{karras2019style} with GRAM-HD, and the results are shown in Figure~\ref{fig:mixing}. By combining styles from the source and target instances in different layers, it is found that styles in shallower layers (layer 1 to 5 in radiance manifold generator) mainly control geometry, while those in deeper layers mainly control appearance. 
Note that our method is not trained with the style mixing strategy.

\subsection{Image embedding and editing}
\begin{figure*}
	\centering
	\includegraphics[height=0.95\textheight]{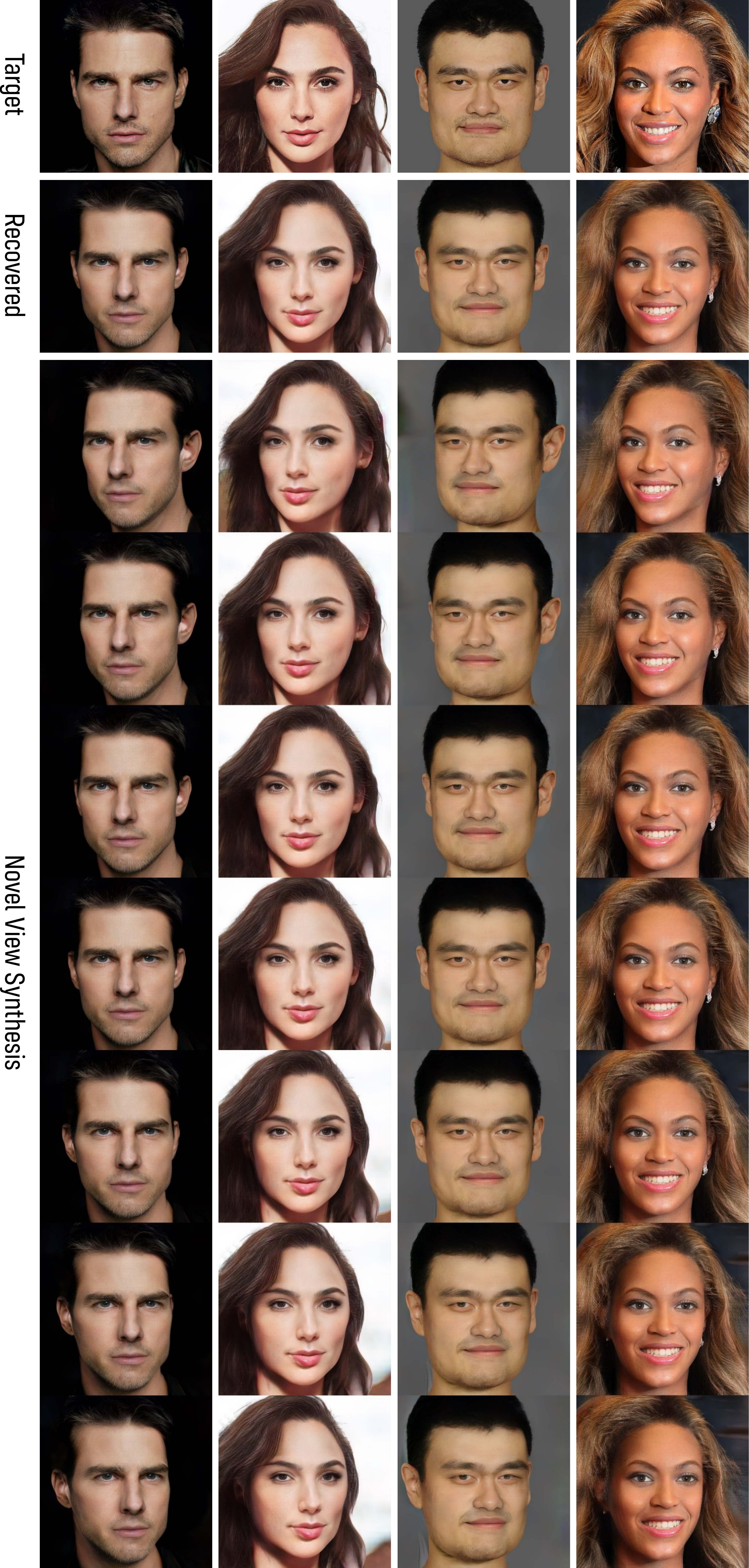}
	\caption{More high-resolution image embedding and editing results.}
	\label{fig:embedding}
\end{figure*}

Figure \ref{fig:embedding} shows more synthesized novel-view images obtained by embedding the given single images. We achieve high-resolution image embedding and pose manipulation with well-maintained 3D consistency even for fine details.

\subsection{Failure cases of generated results}

\begin{figure*}
	\centering
	\includegraphics[width=0.7\textwidth]{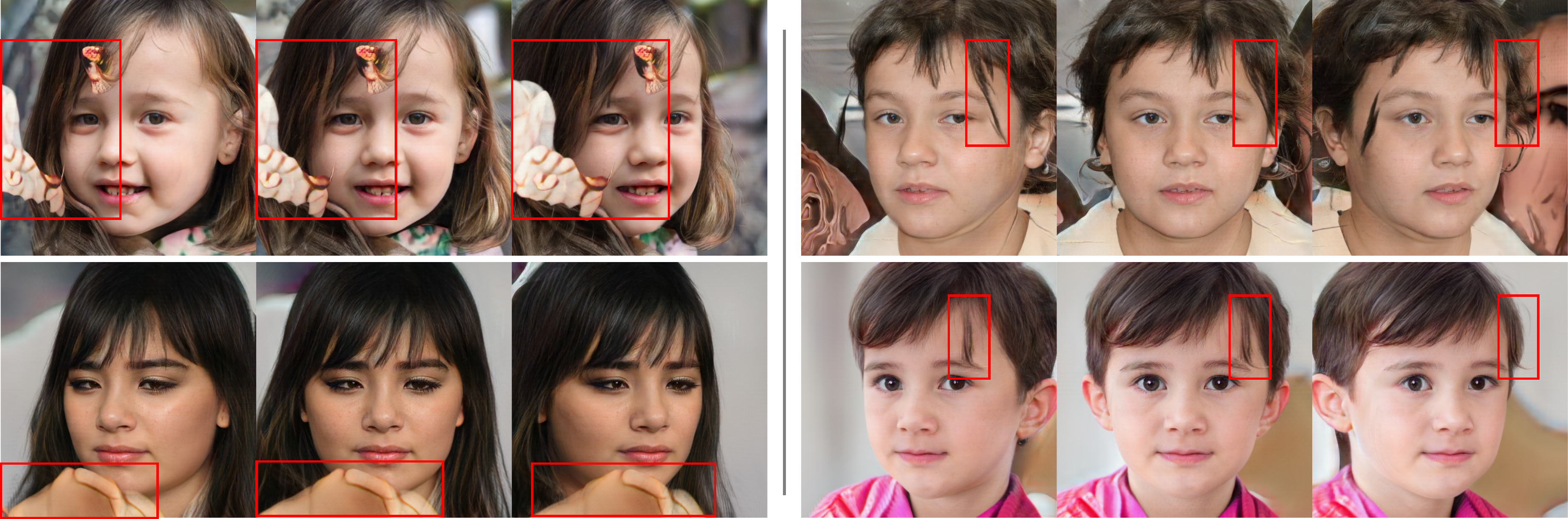}
	\caption{Failure cases. \textbf{Left:} Unwanted floaters on the generation results caused by LR generation (not the super-resolution module). \textbf{Right:} Exaggerated parallax  on some generated instances.}
	\label{fig:failure_case}
\end{figure*}

\paragraph{Floaters}
On some randomly generated results, there could be unwanted floaters in the front, as shown in Figure~\ref{fig:failure_case} (left). These floaters are not produced by super-resolution module, but already exist on the LR radiance manifolds. The reason may be  that the supervision in LR image-space cannot eliminate such floaters for they look fine at low resolution. Jointly training the whole model in one stage may solve the problem, which we leave as our future work.

\paragraph{Exaggerated parallax artifacts}
When rotating the camera, some contents (e.g. hair fringes) on certain generated
instances could be floating at unexpected positions, as shown in Figure \ref{fig:failure_case} (right). We reckon that this is due to the shared surface manifolds across the whole category cannot provide accurate position for all structures of all instances. It could be alleviated by using more shared surface manifolds or learning instance-specific manifolds, which we will also explore in future.

\end{document}